\documentclass[letterpaper, 10 pt, journal, twoside]{IEEEtran} 

\IEEEoverridecommandlockouts                              
\usepackage{times}

\usepackage[numbers]{natbib}
\usepackage{multicol}
\usepackage[hidelinks=true, bookmarks=true]{hyperref}
\usepackage{graphics} 
\usepackage{times} 
\usepackage{amsmath} 
\usepackage{amssymb}  
\usepackage{graphicx}
\usepackage{algorithm}
\usepackage{color}
\usepackage{subcaption}
\usepackage{balance}
\usepackage{colortbl}
\usepackage{threeparttable}
\usepackage{xcolor}
\usepackage{fancyhdr}
\usepackage{tikz}
\usepackage[noend]{algpseudocode}
\usepackage{glossaries}
\usepackage{multirow} 

\usepackage{booktabs}

\def\secref#1{Sec.~\ref{#1}}
\def\figref#1{Fig.~\ref{#1}}
\def\tabref#1{Tab.~\ref{#1}}
\def\eqref#1{Eq.~(\ref{#1})}
\def\algref#1{Alg.~\ref{#1}}

\def\ie{{i.e.}}

\def\etal{\emph{et al.}}
\def\lidar{LiDAR}

\usepackage{amsopn}

\newcommand{\bv}{\mathbf{v}}
\newcommand{\bw}{\mathbf{w}}
\newcommand{\bW}{\mathbf{W}}

\newcommand{\bt}{\mathbf{t}}

\newcommand{\bH}{\mathbf{H}}

\newcommand{\bX}{\mathbf{X}}

\newcommand{\bR}{\mathbf{R}}

\newcommand{\bmu}{\pmb{\mu}}
\newcommand{\bomega}{\pmb{\omega}}

\newcommand{\bJ}{\mathbf{J}}

\newcommand{\cM}{\mathcal{M}}

\newcommand{\bbR}{\mathbb{R}}
\newcommand{\bbSE}{\mathbb{SE}}
\newcommand{\bbSO}{\mathbb{SO}}

\newcommand{\lieSE}{\mathfrak{se}}

\newcommand{\bb}{\mathbf{b}}
\newcommand{\bc}{\mathbf{c}}
\newcommand{\bd}{\mathbf{d}}
\newcommand{\be}{\mathbf{e}}

\newcommand{\bn}{\mathbf{n}}

\newcommand{\bSigma}{\mathbf{\Sigma}}

\newcommand{\bLambda}{\mathbf{\Lambda}}

\def\g2o{$g^2o$}
\def\t2v{\mathrm{log}}
\def\v2t{\mathrm{exp}}
\def\ev2t{\mathrm{ev2t}}

\usepackage{glossaries-extra}
\setabbreviationstyle[acronym]{long-short}
\glssetcategoryattribute{acronym}{nohyperfirst}{true}

\makeglossaries

\newacronym{slam}{SLAM}{Simultaneous Localization and Mapping}
\newacronym{sfm}{SfM}{Structure from Motion}
\newacronym{pgo}{PGO}{Pose-Graph Optimization}
\newacronym{vpr}{VPR}{Visual Place Recognition}
\newacronym{sgd}{SGD}{Stochastic Gradient Descent}
\newacronym{ils}{ILS}{Iterative Least-Squares}
\newacronym{icp}{ICP}{Iterative Closest Point}
\newacronym{gn}{GN}{Gauss-Newton}
\newacronym{lm}{LM}{Levenberg-Marquardt}
\newacronym{pcg}{PCG}{Preconditioned Conjugate Gradient}
\newacronym{map}{MAP}{Maximum-A-Posteriori}
\newacronym{gf}{GF}{Gaussian Filters}
\newacronym{pf}{PF}{Particle Filters}
\newacronym{sdp}{SDP}{Semi-Definite Programming}
\newacronym{bst}{BST}{Binary Search Tree}
\newacronym{ndt}{NDT}{Normal Distributed Transform}
\newacronym{ba}{BA}{Bundle Adjustment}
\newacronym{pca}{PCA}{Principal Component Analysis}
\newacronym{lo}{LO}{LiDAR odometry}
\newacronym{agv}{AGV}{Autonomous Ground Vehicle}

\begin{document}

\title{\LARGE \bf MAD-ICP: It Is All About Matching Data -- Robust and Informed LiDAR Odometry}
\author{\large Simone Ferrari \quad Luca Di Giammarino \quad Leonardo Brizi \quad Giorgio Grisetti
\thanks{All authors are with the Department of Computer, Control, and Management Engineering ``Antonio Ruberti", Sapienza University of Rome, Italy,
Email:\,\,{\tt\footnotesize{\{s.ferrari, digiammarino, brizi, grisetti\}@diag.uniroma1.it.}}}%
}



%

\maketitle

\begin{abstract}
\lidar{} odometry is the task of estimating the ego-motion of the sensor from sequential laser scans. This problem has been addressed by the community for more than two decades, and many effective solutions are available nowadays. Most of these systems implicitly rely on assumptions about the operating environment, the sensor used, and motion pattern. When these assumptions are violated, several well-known systems tend to
perform poorly.

This paper presents a \lidar{} odometry system that can overcome these limitations and operate well under different operating conditions while achieving performance comparable with domain-specific methods. Our algorithm follows the well-known ICP paradigm that leverages a PCA-based kd-tree implementation that is used to extract structural information about the clouds being registered and to compute the minimization metric for the alignment. The drift is bound by managing the local map based on the estimated uncertainty of the tracked pose. To benefit the community, we release an open-source C++ anytime real-time implementation at \url{https://github.com/rvp-group/mad-icp}.
\end{abstract}


\section{Introduction}
\label{sec:intro}
\gls{lo} involves continually estimating the motion of a moving laser scanner within its environment. It is a fundamental component in various robotic applications, including autonomous vehicles and survey systems. Over the past decade, 3D laser scanners have become more affordable while increasing measurement accuracy and resolution.

A reliable \gls{lo} system is an essential building block of more complex applications since it supports fundamental skills such as estimating a model of the environment and interacting with the operating scene. With \gls{lo} becoming a pivotal element in more sophisticated systems, recent advancements emphasize qualities like robustness, simplicity, and ease of use. \textit{Robustness} implies accurate functionality across diverse operating conditions without losing track, encompassing various environments, motion profiles, and \lidar{} types. \textit{Simplicity} and ease of use pertain to both the algorithm and the number of parameters involved.

\begin{figure}[t]
  \centering
  \includegraphics[width=0.99\columnwidth]{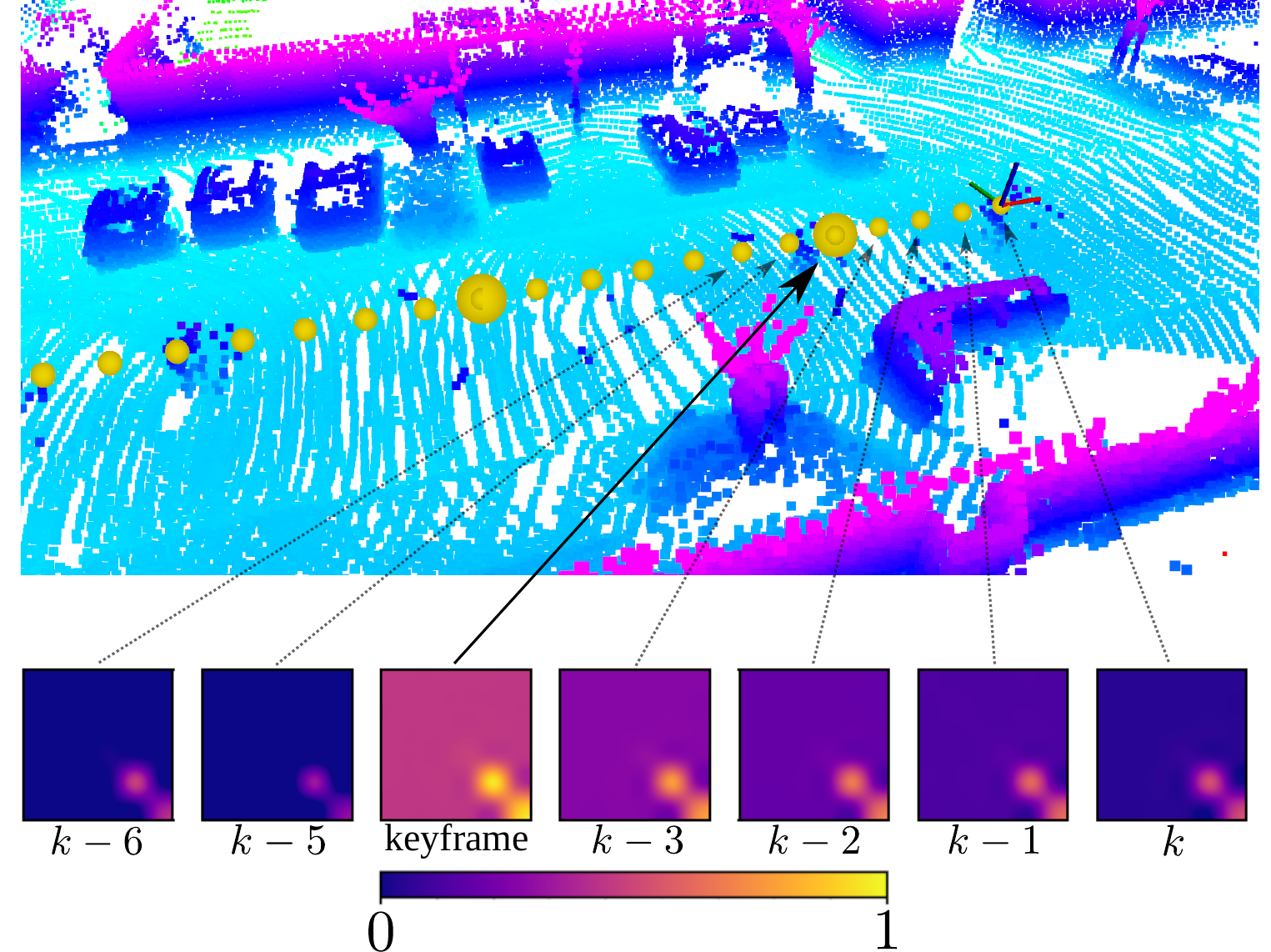}
  \caption{Above is the point cloud generated by our \lidar{} odometry processing a sequence of KITTI \cite{geiger2012we} dataset. The small yellow spheres represent the registered poses corresponding to each sensor measurement. The larger ones are poses related to the point clouds which form the model. Below, the normalized information $6\times6$ matrices are used to select the keyframe poses. The point cloud relative to the pose with the highest information is selected to be part of the model.}
  \label{fig:motivation}
\end{figure}

This paper introduces a novel \gls{lo} method leveraging an efficient and versatile kd-tree data structure and estimated pose uncertainty to dynamically maintain a robust environment model. The tree recursively divides data points based on \gls{pca} \cite{MACKIEWICZ1993303}, reducing a point cloud into leaves representing small planar patches with surface normals. The kd-tree supports isometric transformations and efficient searches with linear and logarithmic worst-case time complexities. The scan-to-model registration is done through \gls{icp} using a \textit{point-to-plane} error metric that compares corresponding leaves, hence performing a data reduction.
The uncertainty of the pose, emerging as a byproduct of the pose estimation, is used to adaptively select which past frame to use to augment the model (\figref{fig:motivation}). This is crucial for accurate and resilient estimations. 

We conducted extensive experiments on a broad range of public data, acquired in heterogeneous environments (urban, highway, vegetation) with different motion profiles (handheld, car-like, drone). The comparative analysis with other SOTA methods reported that our approach has accuracy comparable to or better than other methods. At the same time, it is the only one that never fails in any sequence. We used the same set of parameters for all datasets acquired with the same \lidar{} model.

We release an open-source C++ any-time implementation that can dynamically trade off the accuracy and run-time based on the available computational resources.

\section{Related Work}
Most of the \gls{lo} systems combine a registration algorithm with a local map management. 
Each time a new scan (frame) becomes available, it is registered against the local map (model) to compute the position of maximum overlap.
Subsequently, the local map is augmented with the new measurement, and updated by removing old information.
The model can consist of either a single aggregated global map, or in the union of different local maps/scans. Prominent methods like LeGO-LOAM \cite{shan2018lego}, MULLS \cite{pan2021mulls} and KISS-ICP \cite{vizzo2023kiss} rely on the former scheme, while other SOTA approaches such as in DLO \cite{chen2022direct} use a set of frames. In contrast, SuMa \cite{behley2018efficient} and MD-SLAM \cite{di2022md} anchor on a single frame as a local map and switch to the next frame when the registration support decreases.

The most prominent registration methods in \gls{lo} rely on the \gls{icp} \cite{besl1992method} paradigm. In its basic form, \gls{icp} iteratively aligns two point clouds by estimating a transformation that minimizes the distance between corresponding entities.
This method originally emerged exploiting points, hence employing \textit{point-to-point} distance as error metric. While accurate, this metric is very sensitive to data association, and wrong matches may lead to wrong registrations. 
For this reason, other approaches adopt different error functions. Chen \etal~\cite{chen1992object} propose to exploit the \textit{point-to-plane} error metric. This metric, which considers the surface information, allows for sliding along planar patches. With sufficiently dense data, this approach is more robust and less prone to errors from incorrect data matching \cite{pomerleau2013comparing}.
 

Methods that operate on raw data do not assume any particular structure in the environment, however when some structure is present (e.g. urban car-like environments), a point cloud can be segmented in geometric features such as sharp edges or planar surfaces. 
LOAM \cite{zhang2014loam} has been the root of a family of approaches leveraging geometric feature extraction. Their major drawback is that they resort to 2D projections, which introduce many parameters that are strictly related to the sensor characteristics. In addition, strong assumptions like ground segmentation \cite{shan2018lego} make estimates degrading in non-planar environment (\ie vertical motion involved).




All previously introduced methods estimate a single pose from each scan, treating point clouds as individual entities. However, unconventional \gls{lo} approaches, such as CT-ICP \cite{dellenbach2022ct} and LoLa-SLAM \cite{karimi2021lola}, focus on modeling a continuous trajectory rather than estimating discrete odometry. These methods additionally estimate velocities of intra-scan motion, incorporating on-the-fly point cloud undistortion (deskewing).

With the notable exception of KISS-ICP, all algorithms mentioned so far require the tuning of several parameters that need manual configuration, making them difficult to adapt to a particular context. 

The data structure used to support the correspondence search plays a crucial role in \gls{lo}. Common choices are voxel grids or kd-trees. KISS-ICP \cite{vizzo2023kiss} performs multiple layers of voxelization, keeping for each voxel only a few points. This has the twofold effect of injecting some isotropic noise in the point distribution and de-biasing the point-to-point metric. CT-ICP \cite{dellenbach2022ct} keeps the centroids of the voxels, hence denoising the cloud through averaging. The downside of decimating the data is the potential loss of support during scan registration, which might hinder robustness.

A kd-tree is a binary tree where each node represents an axis-aligned hyperplane that splits the data points into two sections. Typically, the splitting node is chosen as the median point along the largest dimension, leading to a balanced space division \cite{friedman1977algorithm}. In mapping applications, where data often is 3-dimensional, comparative studies show that kd-trees perform excellently in solving k-nearest neighbor problems \cite{elseberg2012comparison}.


Aligning the normal of the splitting plane with the direction of maximum spread computed through \gls{pca} according to \cite{mcnames2001fast} is the optimal choice since it leads to lower the \textit{depth}. 
Differently, common kd-tree implementations such as FLANN \cite{muja2009flann} avoid the \gls{pca} computational burden by using planes aligned with the axes of the initial reference system. This trades off computational complexity for the accuracy of the partitioning but has a negative impact on the correctness and completeness of the search. 
One downside of kd-tree is that each time a new point is added to the representation, the whole tree needs to be changed. FAST-LIO2 \cite{xu2022fast}, addresses the problem of incrementally updating an axis-aligned kd-tree when points are added, removed or re-inserted with small computational overhead, leveraging on concepts similar to the ones of RB-trees \cite{okasaki1999red} used to maintain sorted sets.

On modern computers performing a 3D \gls{pca} has a negligible cost, hence it does not make sense to drop the optimal space partitioning for computation. Furthermore, local surface characteristics such as normal (as the axis of minimal variation) are a natural outcome of \gls{pca} when computed on a densely distributed set of points. 

Relying on these considerations, we designed a \gls{pca} kd-tree specifically for \gls{icp}, supporting isometric transformations and accurate normal extraction.
We represent the local map as a forest of kd-trees, one per keyframe. The keyframes are selected among the most recent scans to maximize the Fisher information of the pose estimated during local map updates.




In summary, the primary contribution of this work is a novel \gls{lo} that provides:
\begin{itemize}
\item a versatile and efficient data structure, capable of encoding all the necessary data in a single representation. This kd-tree is designed to facilitate every required operation, such as plane segmentation, incremental data association, and local map management;
\item an information-aware criterion for adaptive local map updates, based on maximizing the pose information with the aim of limiting the noise injected in the local map;
\item a robust anytime real-time C++ implementation that never loses track among different environments, without requiring parameter tuning.
\end{itemize}

\section{MAD-ICP}
\label{sec:main}
Our pipeline builds on a minimal set of components. The main idea of our technique is to employ a kd-tree for all relevant operations in \gls{lo}.
For each new cloud $\mathcal{C}_k$ delivered by the \lidar, a kd-tree $\mathcal{T}_k$ is built. The result of this preprocess is a data structure that encodes plane segmentation of the cloud (\secref{sec:segmentation}), and allows nearest-neighbor queries.
Each leaf of $\mathcal{T}_k$ will contain a small subset of the point cloud. The task is to match all the terminal nodes against the local map, until the whole tree (representing $\mathcal{C}_k$) is registered. This process is carried out by a point-to-plane \gls{icp} with an incremental data association strategy described in \secref{sec:icp}. The uniqueness of our kd-tree relies on the possibility of transforming the entire data structure without altering the search capabilities, thus avoiding rebuilding after registration. The transformed kd-tree is maintained in a local map, which, thanks to the information aware update criterion, always remains reliable (\secref{sec:local-info}). Finally, velocities are estimated for point cloud undistortion and to provide a good initial guess for \gls{icp} (\secref{sec:velocity}). The full algorithm is outlined in \algref{proc:mad}.

\begin{algorithm}[H]
\small
\caption{MAD-ICP}
\label{proc:mad}
\textbf{local}: local map $\mathcal{M}$, keyframe candidates $\mathcal{Q}$, past odometry poses $\mathcal{P}$, translational velocity $\bv_{k-1}$, rotational velocity $\bomega_{k-1}$\\
\textbf{input}: point cloud $\mathcal{C}_k$
\begin{algorithmic}
	
\If{$k = 1$} \Comment{initialization}
\State $\mathcal{T}_k$ $\leftarrow$ kd-tree($\mathcal{C}_k$),~$\mathcal{M}$ $\leftarrow$ \{$\mathcal{T}_k$\}
\State $P$ $\leftarrow$ \{\textbf{I\textsubscript{4x4}}\},~$\bv_{k-1}$ $\leftarrow$ \textbf{0\textsubscript{3x1}},~$\bomega_{k-1}$ $\leftarrow$ \textbf{0\textsubscript{3x1}}
\State \Return{}
\EndIf
\\
\State $\mathcal{C}_k$ $\leftarrow$ deskew($\mathcal{C}_k$, $\bv_{k-1}$, $\bomega_{k-1}$) \Comment{preprocess}
\State $\mathcal{T}_k$ $\leftarrow$ kd-tree($\mathcal{C}_k$)
\\
\State $^w\bX_{k,\mathrm{guess}}$ $\leftarrow$ integration($^w\bX_{k-1}$, $\bv_{k-1}$, $\bomega_{k-1}$) \Comment{registration}
\State $^w\bX_k$ $\leftarrow$ ICP($\mathcal{M}$, $\mathcal{T}_k$, $^w\bX_{k,\mathrm{guess}}$)
\\
\State $\mathcal{T}_k$.transform($^w\bX_k$) \Comment{tree transformation}
\State $\mathcal{Q}$.push($\mathcal{T}_k$)
\State $\mathcal{P}$.append($^w\bX_k$) \Comment{velocity estimation}
\State $\bv_k$, $\bomega_k$ $\leftarrow$ estimateVelocity($\mathcal{P}$)
\\
\State $p$ $\leftarrow$ $T_k$.leavesPercentage() \Comment{local map update}
\If{$p < p_{\mathrm{th}}$}
\State $M$.update($\mathcal{Q}$)
\EndIf
\Return{}	
\end{algorithmic}
\end{algorithm}

\subsection{Point cloud surface normal segmentation}
\label{sec:segmentation}
\begin{figure}
  \centering
 \includegraphics[width=0.70\columnwidth]{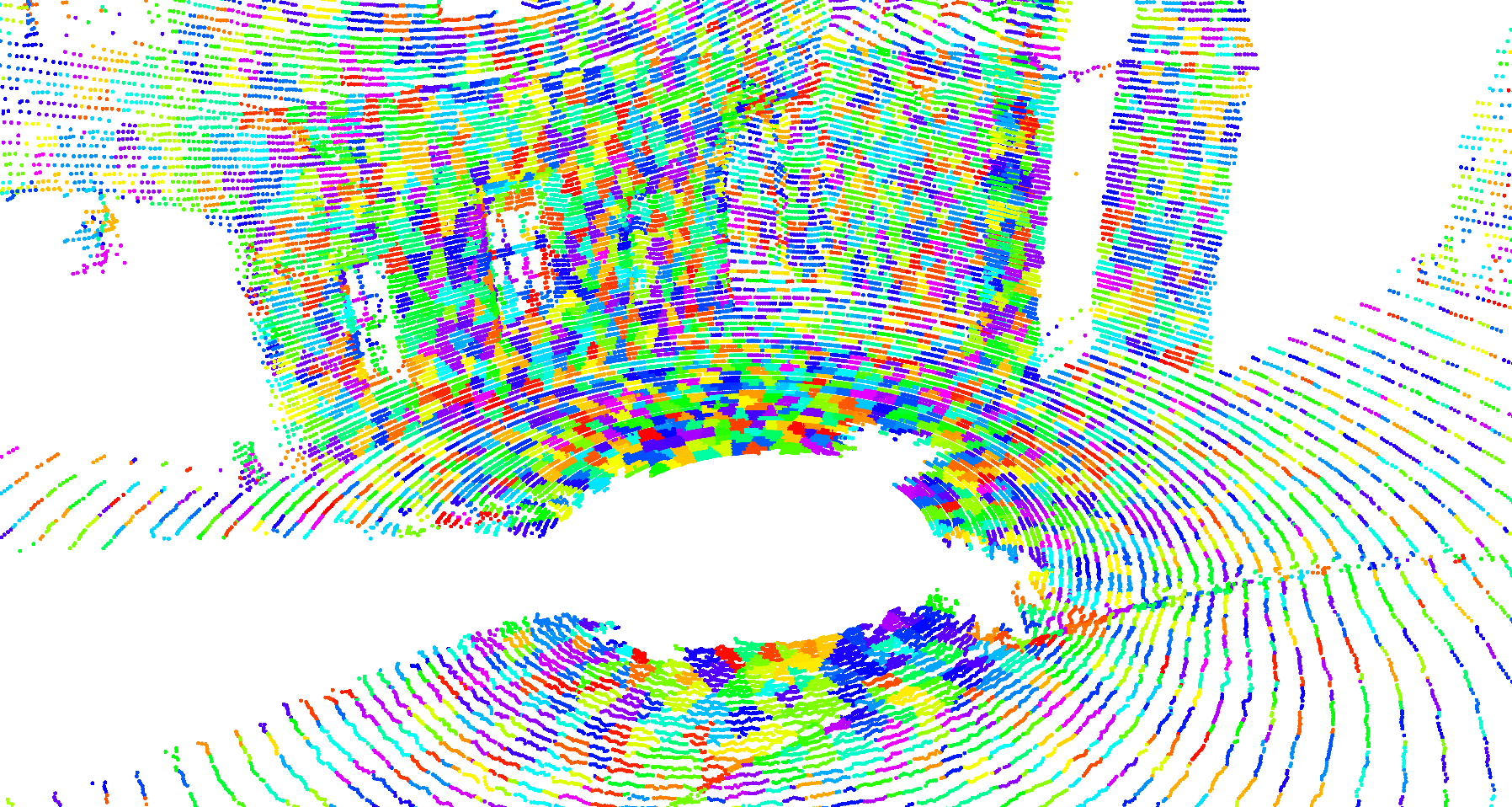}
  \caption{Plane segmentation of a point cloud with our kd-tree explained in \secref{sec:segmentation}. Every leaf is depicted with a different color.}
  \label{fig:leaves}
  \vspace{-0.5cm}
\end{figure} 

Each input cloud $\mathcal{C}=\bc_{1:N}$ is converted into a kd-tree
$\mathcal{T} = \langle \bmu, \bn, \bd, \mathcal{T}_l, \mathcal{T}_r \rangle$. Here $\bmu$ is the mean of the points, $\bn$ is the surface normal and $\bd$ is the direction of maximum spread.
$\mathcal{T}_l$ and $\mathcal{T}_r$ represent respectively the left and right children of the node, and are set to $\emptyset$ for the leaves.

The construction procedure is recursive, and works as follows:
\begin{itemize} 
\item compute mean $\bmu$ and covariance $\bLambda$ from the cloud $\mathcal{C}$;
\item compute the eigenvectors $\bW = \begin{bmatrix} \bw_0 & \bw_1 & \bw_2 \end{bmatrix}$ of $\bLambda$;
\item set the direction of maximum variation as the eigenvector corresponding to the largest eigenvalue $\bd=\bw_2$, and the direction of the normal as the eigenvector associated with the shortest eigenvalue $\bn=\bw_0$;
\item compute the oriented bounding box $\langle b_0, b_1, b_2 \rangle$ by applying the rotation $\bW$ to the cloud $\mathcal C$, and scanning for minimum and maximum along the three axes;
\item if the largest dimension $b_2$ of the bounding box is lower than a threshold $b_{\mathrm{max}}=0.2$ meters, it means that the points are concentrated in the space; hence recursion is stopped;
\item in the case instead the node is not a leaf and spreads over a vast region ($b_2 > b_\mathrm{max}$), the points $\mathcal{C}$ are split into two sets to the left and the right of the splitting plane, based on the predicate: 
\begin{equation}
\bd^T(\bc_i-\bmu)>0.
\label{eq:pred}
\end{equation}
\end{itemize}
The algorithm continues by recursing on these two sets $\mathcal{C}_l$ and $\mathcal{C}_r$, to compute the left and right children $\mathcal{T}_l$ and $\mathcal{T}_r$.

An example of the final plane segmentation can be seen in \figref{fig:leaves}.
This procedure leads to accurate surface normals where the \lidar~pattern is dense enough. Differently, in sparser regions of the cloud, leaves may be defined by only a few noisy points, making their normals unreliable. To address this, in the tree-building process, the normal of a node is propagated to its children when it is sufficiently flat, namely when the length of its shortest dimension is lower than the parameter $b_{\mathrm{min}}$.
This technique allows us to estimate surface normals in most cases without adding further computational complexity to the tree construction. In our experiments, we fix $b_{\mathrm{min}} = 0.1$ meters.
For the pseudocode of the kd-tree construction, we refer the reader to \algref{proc:kd-tree} and \algref{proc:kd-node} in the supplementary material. 


\subsection{Data association and \gls{icp} estimate}
\label{sec:icp}
Our alignment procedure follows an \gls{icp} schema that seeks a transformation $\bX \in \bbSE(3)$ resulting in the best alignment between a reference model $\mathcal M$ and the current measurement.

In our system, the model consists of a forest of kd-trees, expressed in the world frame $w$, while the measurement at time $k$ is represented by the kd-tree $\mathcal{T}_k$, expressed in \lidar{} frame.

\gls{icp} alternates two steps: data matching and optimization. During data association, each leaf of $\mathcal{T}_k$ is transformed in $w$, based on the current transformation estimate, and its nearest neighbor in $\mathcal M$ is found.
Let $^w\bX_k$ be the sensor pose at instant $k$
expressed in the world frame $w$ and let $q = \langle \bmu_q, \bn_q, \bd_q, \emptyset, \emptyset \rangle$ be a generic leaf in $\mathcal{T}_k$.
The search within a tree $\mathcal{T} \in \mathcal{M}$ is a recursive procedure that follows the same predicate of \eqref{eq:pred} used during the tree-building process:
\begin{equation}
\bd ^T (^w\bX_k \bmu_q - \bmu)>0. \label{eq:predicate}
\end{equation}
The search starts from the root of each tree in the model and iteratively continues by traversing the child indicated by \eqref{eq:predicate}. The recursion stops when a leaf $l = \langle \bmu_l, \bn_l, \bd_l, \emptyset, \emptyset \rangle$ is reached, thus $q$ and $l$ form a patching pair $\langle l, q \rangle$.

At this point, it is necessary to eliminate outliers. This is accomplished by verifying if the Euclidean distance between $\bmu_q$ and $\bmu_l$ is greater than the search radius $r$, which is computed as:

\begin{equation}
r = b_\mathrm{max} + \lVert \bmu_q\rVert~b_{\mathrm{ratio}}.
\end{equation}

Here, $b_{\mathrm{max}}$ denotes the maximum size of the leaves, and $b_{\mathrm{ratio}}$ represents a parameter that increases the search size radius as the norm of the query $\bmu_q$ grows. 

Generally, 3D \gls{icp} seeks to estimate a transformation matrix $\bX \in \bbSE(3)$ composed by a rotation matrix $\bR \in \bbSO(3)$ and translation vector $\bt \in \bbR^3$. In our case, we estimate the isometry $^w\bX_k$ denoting the sensor pose at instant $k$ expressed in the world frame $w$. This is computed by iteratively minimizing a point-to-plane error term using Gauss-Newton. Given a match $\langle l, q \rangle$, the point-to-plane error is computed as:
\begin{equation}
e = \bn_l^T\left(^w\bX_k~\bmu_q - \bmu_l \right).
\label{eq:cost}
\end{equation}
This shows the difference between $^w\bmu_q$ and $\bmu_l$, projected onto the normal $\bn_l$. In addition, we employ the Huber robust loss and in our experiments we set its weighting parameter to $\rho_\mathrm{ker} = 0.1$.

Upon convergence, a Gaussian covariance approximation of the solution can be found according to \cite{censi2007accurate}. More specifically, since our approach carries on the optimization of the Lie Algebra $\lieSE(3)$, the inverse covariance of the estimate $\bSigma^{-1} \in \bbR^{6 \times 6}$ is the system matrix of the Gauss-Newton algorithm.

Thanks to the density offered by modern \lidar{}, the point-to-plane error exhibits robust performances, steering the optimization process toward explicit surfaces defined in the local map.

Our versatile kd-tree composed of splitting planes and surface normals can be directly transformed by the current estimate. Specifically, this happens multiple times: during optimization, where leaves are transformed over iterations to carry on the next correspondence search, and at the end, when the whole kd-tree $\mathcal{T}_k$, with its splitting planes, is transformed in the world frame through the final estimate $^w\bX_k$. Tree transformation does not alter data association capabilities and is performed with limited computational effort. Finally, the transformed kd-tree is pushed into a queue $\mathcal{Q}$ of candidate frames for local map updates.

The pseudocode is available in \algref{proc:icp}, in the supplementary material.



\subsection{Local map representation and update}
\label{sec:local-info}
Our local map consists of a forest $\mathcal{M}$ of kd-trees, each of them corresponding to a keyframe.
This model, composed of independent kd-trees, preserves the accuracy and reliability of each point cloud, leaving the surface normals unchanged. Moreover, compared to either incremental kd-trees or voxel grids, our approach is simpler to implement because updating the local map is accomplished by merely pushing a new tree into it.

Keeping a reliable representation of the environment is crucial for effective registration. Many works in this field update the local map with every new observation \cite{shan2018lego, pan2021mulls, vizzo2023kiss, dellenbach2022ct}. Instead, we believe that updating only when needed saves us from continuously injecting noise into our model. Furthermore, our local map update strategy is information aware, meaning that only the least uncertainty match is considered, avoiding inserting in $\mathcal{M}$ trees that are poorly matched during the optimization phase.

In principle, the local map update could be triggered by a pose change or by a change in the observed space. The first option is challenging to tune and lacks generalization, as it depends on the specific characteristics of the \lidar~sensor. For instance, sensors with an omnidirectional FoV should not trigger a local map update if they are rotated, as the portion of the observed space remains invariant to rotation.
For this reason, we trigger local map update when $p < p_\mathrm{th}$, where $p$ is the percentage of leaves in $\mathcal{T}_k$ that matched with $\mathcal{M}$, and $p_\mathrm{th}$ a parameter that we set to 0.8.
 
When this happens the frame with the lowest uncertainty $\bSigma$ (presented in \secref{sec:icp}) is extracted from the list of frame candidates $\mathcal{Q}$. In order to compare covariance matrices associated with different candidates, several functions — known as \textit{optimality criteria} — have been proposed \cite{placed2023survey}; in our algorithm, we employ the matrix determinant also known as \textit{D}-optimality criterion \cite{wald1943efficient}. This candidate becomes our new keyframe and is pushed inside the local map $\mathcal{M}$.

The determinant of poses uncertainty is considered also in \cite{kuo2020redesigning}, but with the aim of triggering a new keyframe selection. In contrast, we perform a minimization over $\bSigma$ for choosing the new keyframe, and we trigger this event based on the support of the current observation.

We need to point out that when revisiting an already short-term explored area, our algorithm behavior degenerates to a localization process, anchoring to previously created keyframes without the need for new updates.

\subsection{Velocity estimation and initial guess refinement}
\label{sec:velocity}
Velocity estimation is demanded for two purposes: first, to compensate for motion distortion by deskewing the current point cloud, and second, to predict the next pose, enabling a good initial guess for the \gls{icp} algorithm. 

The full point cloud acquisition of a \lidar{} is rather continuous. For this reason, motion during data capture reflects in point cloud distortion, leading to misaligned points, given the scanning frequency of the sensor. Motion compensation is essential to correct this effect, ensuring accurate spatial representation in the 3D environment \cite{salem2023check}. As in \cite{anderson2013ransac}, given the high rate at which each fragment of a point cloud is acquired, for deskewing, we employ a classic constant motion model.

Differently, given the lower rate at which each point cloud is fully measured, we adopt a smoothing technique in order to provide a good initial guess to our odometry estimation process. Our solution involves computing velocity using Gauss-Newton as in \gls{icp} (\secref{sec:icp}). We smooth the translational and rotational velocities, respectively defined as $\bv_k \in \bbR^3$ and $\bomega_k \in \bbR^3$ for the current time $k$ using the last $n$ poses. In our experiments, we set $n=10$. Specifically, given the transformation $^i\bX_k$ from the current time instant $k$ to a previous one $i$ (with $\Delta t_i = k - i$), we compute the errors as follows:
\begin{align}
	\be_{\bv} &= \Delta t_i~ \bv_k - ^i\bt_k, \\
	\be_{\bomega} &= \Delta t_i~ \bomega_k - \mathrm{Log}(^i\bR_k),
\end{align}
where the operator $\mathrm{Log}(\cdot) \in \bbR^3$ is the logarithmic map of the rotation matrix evaluated at the identity.

\begin{figure*}[t]
  \centering
 \includegraphics[width=1\textwidth]{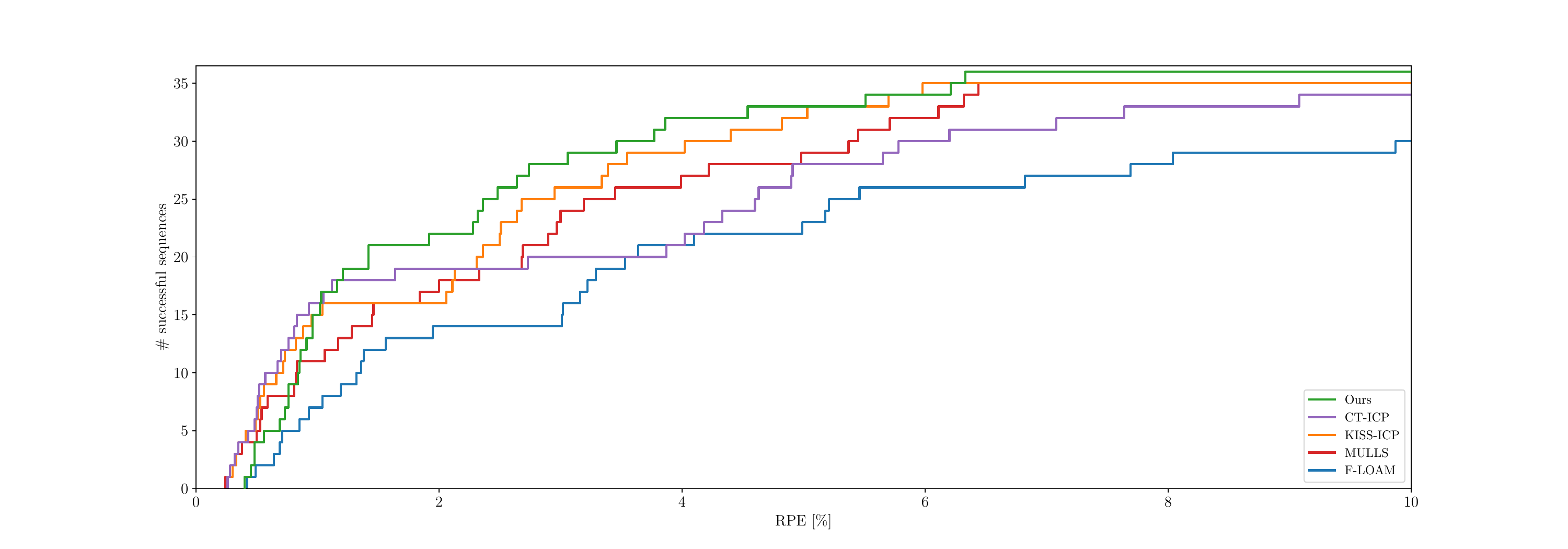}
  \caption{The plot shows the cumulative error curve for each of the compared approaches, computed using all the sequences RPE [\%] of \tabref{tab:results}. For a given error value on the horizontal axis, the vertical axis shows how many sequences a method achieves a lower error.}
  \label{fig:rpe_cumulative}
  \vspace{-0.7cm}
\end{figure*}

\section{Experimental Evaluation}
\label{sec:exp}
In this section, we report the results of our method extensively tested on several public datasets. In order to quantitatively compare our algorithm with other SOTA approaches, we employ the KITTI benchmark \cite{geiger2012we} and the VBR benchmark \cite{bg2024vbr} to asses, respectively, local accuracy and robustness of odometry estimates (\secref{sec:evalsetup}), exploiting the Relative Pose Error (RPE). We conducted all the experiments on a machine with CPU i9-13900kf with 16 physical cores. Our system consists of a minimal and robust \gls{lo} accompanied by a single set of parameters (\secref{sec:parameters}). Our quantitative results, presented in \secref{sec:comparison}, show that our \gls{lo} performs on par or better than other SOTA approaches. Experimental evaluation supports our claim in a wide range of heterogeneous environments, including highways, narrow buildings, stairs, static and dynamic settings. We also consider various \lidar~sensors and arbitrary motion profiles (encompassing different types of robots, cars, and handheld devices).

\begin{figure}[t]
  \centering
 \includegraphics[width=0.99\columnwidth]{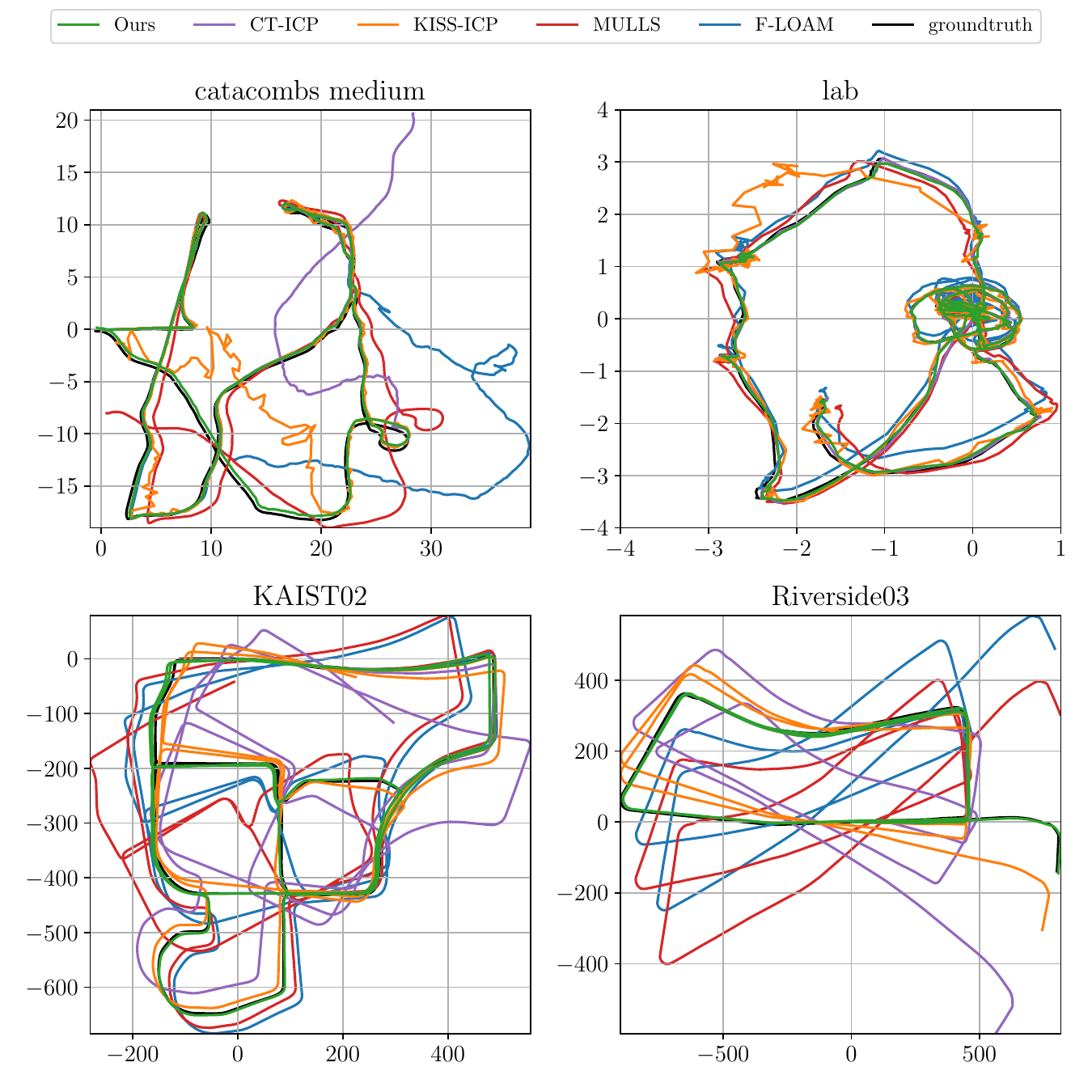}
  \caption{The plots illustrate the qualitative absolute consistency of the compared approaches within some sequences of the employed datasets.}
  \label{fig:plots}
\end{figure}

We report an extensive experimental campaign on the following datasets: KITTI \cite{geiger2012we}, Mulran \cite{kim2020mulran}, Newer College \cite{ramezani2020newer}, Hilti 2021 \cite{helmberger2022hilti}.
This experimental evaluation includes the comparison with four other \lidar~odometries released as open-source: F-LOAM \cite{floam2016wang}, MULLS \cite{pan2021mulls}, CT-ICP \cite{dellenbach2022ct} and KISS-ICP \cite{vizzo2023kiss}. 
Our quantitative results are complemented by qualitative plots for global consistency of the compared trajectories (\figref{fig:plots}). \tabref{tab:iaware} quantifies the enhancements of our information-aware strategy compared with a naive selection. For runtimes analysis of our system (\figref{fig:accvst}, \figref{fig:profiling}), further qualitative comparisons (\figref{fig:plots_2}, \figref{fig:overall}), and tests with RBG-D data (\figref{fig:eth}), we refer the reader to our supplementary material.


\subsection{Parameters}
\label{sec:parameters}
Our approach relies on a set of six parameters that do not require tuning when transitioning between different \lidar~sensors or environments. This configuration demonstrates versatility, as it performs consistently across various scenarios. The parameters are already presented over \secref{sec:main}, but we compactly list them here for ease of reading:
\begin{itemize}
	\item $b_{\mathrm{max}} = 0.2$: the maximum size (in meters) of kd-tree leaves;
	\item $b_{\mathrm{min}} = 0.1$: (in meters) in the tree-building process, when a node is flatter than this parameter, its normal is propagated and assigned to its children;
	\item $b_{\mathrm{ratio}} = 0.02$: the increase factor of the search radius $r$ needed in the data association;
	\item $p_{\mathrm{th}} = 0.8$: the threshold needed to trigger the update of the local map $M$, ensuring an update when the current point cloud is registered for less than 80\%;
	\item $\rho_{\mathrm{ker}} = 0.1$: a parameter of the Huber robust estimator;
	\item $n = 10$: the number of last poses used to estimate the smoothed velocity.
\end{itemize}

\subsection{Evaluation setup}
\label{sec:evalsetup}
For evaluating the accuracy, we employed the most common and widely used ranking method, namely the KITTI benchmark \cite{geiger2012we}, which involves computing the Relative Pose Error (RPE) over a set of subsequences. The length of these subsequences is set to (100, 200, 300, ..., 800) meters for long trajectories like those in KITTI \cite{geiger2012we}, Mulran \cite{kim2020mulran} and Newer College with OS1 \cite{ramezani2020newer}. Instead, for relatively short recordings like Newer College with OS0 and Hilti 2021 \cite{helmberger2022hilti} we opted to reduce these lengths to (10, 20, 30, ..., 80) meters.
The evaluation ranks methods based on the average of these error values, measured in percentage.
The average of each dataset is computed over all the errors over all possible subsequences, while the final score at the end of the table is computed simply as the mean of the average of each dataset. It is important to note that the total error reported here for each pipeline includes only the sequences where the \gls{lo} does not fail, ensuring a fair comparison.

For evaluating the robustness, we employed the VBR benchmark \cite{bg2024vbr}, which consists of computing a cumulative error curve from all the sequences RPE [\%]. The method ranking is determined as the area under the curve up to 10 \%, so the larger the better. This metric rewards the robustness of evaluated methods, since a successful result on a sequence usually adds much more area under the curve than slightly improving the accuracy on many sequences.
\begin{table*}[t]
\centering
\scalebox{0.8}{
\begin{tabular}{cccccccccc}
\cline{2-10}
                              & catacombs easy & catacombs medium & cloister      & math easy     & math medium   & quad easy     & quad medium   & stairs        & avg           \\ \hline
\multicolumn{1}{c}{Ours} & \textbf{1.16}  & \textbf{1.42}    & \textbf{1.42} & \textbf{0.40} & \textbf{0.56} & \textbf{2.64} & \textbf{5.51} & \textbf{0.91} & \textbf{1.84} \\ \hline
\multicolumn{1}{c}{Ours w/o IA}    & 1.31           & 2.13             & 1.45          & 0.45          & 0.79          & \textbf{2.64} & \textbf{5.51} & 1.06          & 1.97         \\ \bottomrule
\end{tabular}}
\caption{Quantitative results obtained from Newer College dataset \cite{ramezani2020newer} to show the impact of our information aware local map update strategy discussed in \secref{sec:local-info}. Applying our algorithm in conjunction with this information-aware strategy improves accuracy. Notably, in sequences characterized by a confined motion within four walls, such as \textit{quad} sequences, there is no discernible difference as the redundant motion renders all information equally effective.}
\label{tab:iaware}
\vspace{-0.5cm}
\end{table*}
\subsection{Comparison with State-of-the-art}
\label{sec:comparison}
For comparison, we explicitly selected the most recent, best-performing SOTA \gls{lo}, appropriately configuring them for specific motion profiles, sensors, and environmental characteristics when needed.

The first work is F-LOAM \cite{floam2016wang}, which aims to decrease the computational burden of previous LOAMs approaches by transforming the iterative processes into a two-stage distortion compensation method. It uses special features, such as edges with higher local smoothness and planar features with lower smoothness, to improve matching, by projecting the cloud into a range image.

A recent approach that instead relies on the basics of \gls{icp} is MULLS (Multi-metric Linear Least Square) \cite{pan2021mulls}. It is designed to be independent of \lidar~specifications, eliminating the need for converting \lidar~data to rings or range images. It extracts geometric feature points, categorizes them (ground, facade, pillars), and performs ego-motion estimation using multi-metric linear Least Squares \gls{icp}. 

CT-ICP \cite{dellenbach2022ct} incorporates the motion compensation into registration by adding intra-scan velocities to the variables being estimated within \gls{icp} and uses a point-to-plane metric. It relies on a consistent voxel-based local map that is updated with the new scan after registration and deskewing.

KISS-ICP \cite{vizzo2023kiss} is another modern approach that leverages \gls{icp}. This pipeline adopts the constant velocity model for motion compensation. New scans are first subsampled and then matched with the local map to find correspondences. This process is accompanied by an adaptive thresholding scheme. Then, \gls{icp} is performed with a point-to-point metric, and the local map is updated as in CT-ICP.

Results using the KITTI benchmark (explained in \secref{sec:evalsetup}) are shown in \tabref{tab:results}. 

The first section of the table is about the eleven training sequences (00-10) of KITTI \cite{geiger2012we}, recorded by car. These first results exhibit only small differences in performance among the five pipelines. This dataset is recorded with a Velodyne HDL-64. However, it is important to note that KITTI is still the most famous vision benchmark, but the inconsistencies in the baselines are on the order of meters \cite{bg2024vbr}, since their ground truth system relies only on RTK-GPS measurements. This ensures good global consistency but not local, which, unfortunately, is required when evaluating odometry systems.

The second section includes the Mulran \cite{kim2020mulran} data collection. This is a car dataset recorded with Ouster OS1-64. In some of these sequences, our method performs slightly better than the others. However, even in this case, the public ground truth trajectories are of poor quality, even worse than KITTI, always relying only on RTK-GPS. Still, they provide some insights into the behavior of these \gls{lo} in large outdoor scenarios.

In the third and fourth sections, we conducted tests on the Newer College \cite{ramezani2020newer} dataset, which is divided into sequences recorded through Ouster OS0-128 and OS1-64 sensors. This collection allowed us to rank the pipelines when dealing with handheld data. This time, the baselines are highly accurate and allow for a fair comparison, given the ground truth processed by registering \lidar{} scans to a high-definition map built using a 3D Leica laser scanner. It is evident that our method, KISS-ICP and CT-ICP perform quite similarly in the OS1 sequences. This result was expected because, despite a change in motion profile, OS1 data still represents large and open environments, similar to those found in outdoor KITTI and Mulran streets.

The situation is entirely different with the OS0 data, where we outperform all the compared methods, demonstrating the versatility and accuracy of our approach. 
Very similar results can be found in the last table section, showing the only sequences of the Hilti 2021 \cite{helmberger2022hilti} dataset where the ground truth is provided. The two sequences are acquired using a quadrotor and handled, respectively, with an Ouster OS0-64.

At the end of the table, we present the average score from all the datasets, indicating that our approach performs slightly better than the compared methods. It is important to note that the total error reported here for each pipeline includes only the sequences where other \gls{lo} do not fail, ensuring a fair comparison.
Finally, \tabref{tab:cumulative} reports for each method the area under the curve depicted in \figref{fig:rpe_cumulative}. Our method achieves the highest ranking. This suggests that our system is not only more accurate on average but also exhibits robustness in challenging scenarios where other current SOTA methods lose track.

\definecolor{Gray}{gray}{0.85}
\begin{table}    
\vspace{0.2cm}
		\centering

            \begin{tabular}{p{0.05cm}!{\color{Gray!40!Gray}\vrule width 1.8pt}p{1.2cm}<{\centering}p{0.9cm}<{\centering}p{0.9cm}<{\centering}p{0.9cm}<{\centering}p{0.9cm}<{\centering}p{0.9cm}<{\centering}}

			\toprule

			\multicolumn{1}{p{0.01cm}}{} & & Ours & KISS-ICP & F-LOAM & MULLS & CT-ICP \\
			\midrule
			\multirow{11}{*}{\rotatebox[origin=c]{90}{KITTI}}			
			&0 & 0.73 & 0.51 & 0.69 & 0.53 & \textbf{0.50} \\
			&1 & 0.85 & 0.72 & 1.95 & 0.81 & \textbf{0.67} \\
			&2 & 0.76 & 0.53 & 1.04 & 0.59 & \textbf{0.51} \\
			&3 & 0.84 & \textbf{0.66} & 1.32 & 1.84 & 0.70 \\
			&4 & 0.69 & \textbf{0.35} & 0.71 & 0.38 & \textbf{0.35} \\
			&5 & 0.48 & 0.30 & 3.64 & 0.32 & \textbf{0.26} \\
			&6 & 0.48 & 0.26 & 0.49 & \textbf{0.24} & 0.28 \\
			&7 & 0.45 & 0.33 & 0.42 & \textbf{0.30} & 0.32 \\
			&8 & 1.21 & 0.82 & 0.93 & 0.83 & \textbf{0.81} \\
			&9 & 1.02 & 0.49 & 0.64 & 0.54 & \textbf{0.48} \\
			&10 & 0.96 & 0.56 & 1.19 & 0.82 & \textbf{0.52} \\
			\arrayrulecolor{Gray}  
			\midrule
			&avg & 0.82 & 0.54 & 1.25 & 0.60 & \textbf{0.53} \\
			\arrayrulecolor{black}  
			\midrule
			\multirow{12}{*}{\rotatebox[origin=c]{90}{Mulran}}			
			& DCC01 & 3.06 & \textbf{2.95} & 4.10 & 3.19 & 4.91 \\
			& DCC02 & \textbf{2.28} & 2.50 & 3.53 & 2.33 & 4.18 \\
			& DCC03 & \textbf{1.92} & 2.11 & 3.16 & 2.90 & 4.63 \\
			& KAIST01 & \textbf{2.36} & \textbf{2.36} & 3.22 & 2.68 & 4.33 \\
			& KAIST02 & 2.32 & \textbf{2.31} & 3.01 & 3.45 & 4.02 \\
			& KAIST03 & 2.74 & \textbf{2.68} & 3.29 & 2.97 & 4.60 \\
			& River.01 & 3.77 & \textbf{3.55} & 5.21 & 4.98 & 7.08 \\
			& River.02 & 3.46 & \textbf{3.34} & 4.99 & 5.45 & 6.20 \\
			& River.03 & \textbf{2.48} & 2.51 & 5.18 & 3.99 & 5.78 \\
			& Sejong01 & 4.54 & \textbf{4.40} & 13.59 & 6.44 & 9.08 \\
			& Sejong02 & 6.21 & \textbf{4.82} & 7.69 & 5.37 & 7.64 \\
			& Sejong03 & 6.33 & 5.03 & 6.82 & 6.32 & \textbf{4.90} \\
			\arrayrulecolor{Gray}  
			\midrule
			& avg & 4.34 & \textbf{3.82} & 6.97 & 4.93 & 6.25 \\
			\arrayrulecolor{black}  
			\midrule
			\multirow{8}{*}{\rotatebox[origin=c]{90}{NC0}}			
			& cat. easy & 1.16 & 2.06 & 1.36 & 2.00 & \textbf{1.12} \\
			& cat. med. & \textbf{1.42} & \textcolor{red}{5.70} & \textcolor{red}{11.02} & 3.00 & \textcolor{red}{10.42} \\
			& cloister & \textbf{1.42} & 3.39 & 5.46 & 1.45 & 1.64 \\
			& m. easy & \textbf{0.40} & 0.41 & 0.85 & 0.50 & 0.43 \\
			& m. med. & \textbf{0.56} & 0.73 & \textcolor{red}{18.05} & 1.06 & 0.57 \\
			& quad easy & \textbf{2.64} & \textbf{2.64} & 3.02 & 2.69 & 2.73 \\
			& quad med. & \textbf{5.51} & 5.98 & \textcolor{red}{31.06} & 5.71 & 5.65 \\
			& stairs & \textbf{0.91} & \textcolor{red}{17903.09} & 8.04 & \textcolor{red}{23.84} & \textcolor{red}{29.69} \\
			\arrayrulecolor{Gray}  
			\midrule
			& avg & \textbf{1.84} & 2.60 & 3.34 & 2.24 & 2.02 \\
			\arrayrulecolor{black}  
			\midrule
			\multirow{3}{*}{\rotatebox[origin=c]{90}{NC1}}			
			& short & 0.86 & 0.88 & 1.56 & 1.28 & \textbf{0.83} \\
			& long & 0.96 & 0.95 & \textcolor{red}{9.87} & 1.17 & \textbf{0.93} \\
			& parkland & \textbf{1.03} & 1.04 & \textcolor{red}{14.79} & \textcolor{red}{6.11} & 1.05 \\
			\arrayrulecolor{Gray}  
			\midrule
			& avg & 0.93 & 0.94 & 1.56 & 1.21 & \textbf{0.90} \\
			\arrayrulecolor{black}  
			\midrule
			\multirow{2}{*}{\rotatebox[origin=c]{90}{Hilti}}			
			& drone & \textbf{3.86} & 4.02 & \textcolor{red}{19.49} & 4.22 & 3.87 \\
			& lab & \textbf{0.76} & 2.13 & 1.38 & 1.46 & \textbf{0.76} \\
			\arrayrulecolor{Gray}  
			\midrule
			& avg & 2.79 & 3.37 & \textbf{1.38} & 3.26 & 2.79 \\
			\arrayrulecolor{black}  
			\midrule & tot avg & \textbf{2.14} & 2.25 & 2.90 & 2.45 & 2.50 \\
			\bottomrule
		\end{tabular}
  \caption{Results obtained using the KITTI benchmark over different public datasets. The table shows the Relative Pose Error (RPE), which is reported in \%. For each dataset we show the average, at the end we show the total average. In order to ensure a fair comparison, failure results for the average calculations (the one highlighted in red) are not included.}
  \label{tab:results}
\end{table}

\begin{table}

\centering
\begin{tabular}{cccccc}
\cline{2-6}
                              & Ours & KISS-ICP & F-LOAM & MULLS & CT-ICP \\ \hline
\multicolumn{1}{c}{AUC} & \textbf{288.57} & 275.98 & 205.24 & 262.08 & 247.67 \\ \bottomrule
\end{tabular}
\caption{Area under the curve for each approach compared in \figref{fig:rpe_cumulative}. The higher, the better.}
\label{tab:cumulative}
\end{table} 

\section{Conclusion}
\label{sec:conclusion}
The key of our \gls{lo} is the clever reuse of previously computed information in many stages: the \gls{pca} used to partition the space and construct the search structure is also used to classify the surface and to assign the normals.
Similarly, the Fisher information matrix, resulting from registration, is used for adaptive local map updates that limit the drift. 
Most of the calculations are controlled by the input data, through the implicit statistics computed along the different stages. This results in a robust, simple, and adaptive \gls{lo} that addresses the limitations of most of the existing solutions overfitted to specific data domains. Our claims are backed up by an extensive set of comparative experiments.
To contribute to the community, we provide an open-source C++ anytime real-time implementation. As future works, to make our system even more robust, we envision an odometry that fuses an inertial sensor in a continuous way. 

\bibliographystyle{IEEEtran}
\bibliography{robots}

\section{Supplementary Material}
In this section we expand the main procedures that were not fully detailed.
\subsection{Algorithms}
\label{sec:algorithm}
In \algref{proc:kd-tree} and \algref{proc:kd-node} we provide a formalization of the kd-tree building process. Additionally, in \algref{proc:icp}, we offer a comprehensive explanation of how this data structure is utilized for identifying matches and estimating the transformation.
 
\begin{algorithm}
    \small
    \caption{kd-tree}
    \label{proc:kd-tree}
    \textbf{input}: point cloud $\mathcal{C}$\\
    \textbf{output}: kd-tree $\mathcal{T}$
    \begin{algorithmic}
        
        \State $\mathcal{T}$ $\leftarrow$ kd-node($\mathcal{C}$,~$false$,~$\textbf{0\textsubscript{3x1}}$)
        
        \State \Return{$\mathcal{T}$}	
    \end{algorithmic}
\end{algorithm}

\begin{algorithm}
    \small
    \caption{kd-node}
    \label{proc:kd-node}
    \textbf{input}: point cloud $\mathcal{C}$, boolean flag $\texttt{inherits\_normal}$, inherited normal $\bn_{p}$\\
    \textbf{output}: node $\mathcal{T}$
    \begin{algorithmic}
        \State $\bmu$, $\bLambda$ $\leftarrow$ meanAndCovariance($\mathcal{C}$) \Comment{PCA}
        \State $\bW$ $\leftarrow$ eigenvectors($\bLambda$)
        \State $\bw_0$, $\bw_1$, $\bw_2$ $\leftarrow$ sortEigenvectors($\bW$)
        \\
        \State $\mathcal{T}$ $\leftarrow$ emptyKdNode() \Comment{assign node attributes}
        \State $\mathcal{T}$.$\bmu$ $\leftarrow$ $\bmu$
        \State $\mathcal{T}$.$\bn$ $\leftarrow$ $\bw_0$
        \State $\mathcal{T}$.$\bd$ $\leftarrow$ $\bw_2$
        \State $\mathcal{T}$.$\bb$ $\leftarrow$ computeBoundingBox($\mathcal{C}$, $\bmu$, $\bW$)
        \\
        \If{$\mathcal{T}.b_2 < b_\mathrm{max}$} \Comment{base case}
        \If{$\texttt{inherits\_normal}$}
        \State $\mathcal{T}$.$\bn$ $\leftarrow$ $\bn_{p}$ \Comment{enhance normal}
        \EndIf
        \Return{$\mathcal{T}$}
        \EndIf
        \\
        \If{\textit{\textbf{NOT}} $\texttt{inherits\_normal}$ \textit{\textbf{AND}} $\mathcal{T}.b_0 < b_\mathrm{min}$}
        \State $\texttt{inherits\_normal}$ $\leftarrow$ $true$
        \State $\bn_{p}$ $\leftarrow$ $\mathcal{T}$.$\bn$ \Comment{propagate normal if node is flat}
        \EndIf
        \\
        \State $\mathcal{C}_l$, $\mathcal{C}_r$ $\leftarrow$ splitCloud($\mathcal{C}$, $\mathcal{T}$.$\bmu$, $\mathcal{T}$.$\bd$) \Comment{recursive case}
        \State $\mathcal{T}$.$\mathcal{T}_l$ $\leftarrow$ kd-node($\mathcal{C}_l$, $\texttt{inherits\_normal}$, $\bn_{p}$)
        \State $\mathcal{T}$.$\mathcal{T}_r$ $\leftarrow$ kd-node($\mathcal{C}_r$, $\texttt{inherits\_normal}$, $\bn_{p}$)
        \\
        \Return{$\mathcal{T}$}	
    \end{algorithmic}
\end{algorithm}

\begin{algorithm}
    \small
    \caption{ICP}
    \label{proc:icp}
    \textbf{input}: local map $\mathcal{M}$, current tree $\mathcal{T}_k$, prediction $^w\bX_{k,\mathrm{guess}}$\\
    \textbf{output}: estimated pose $^w\bX_k$\\
    \begin{algorithmic}
        
        \State $^w\bX_k$ $\leftarrow$ $^w\bX_{k,\mathrm{guess}}$ \Comment{initialization}\\
        
        \While{timeNotElapsed()} \Comment{anytime ICP rounds}
        \State $\bH$ $\leftarrow$ $\textbf{0\textsubscript{6x6}}$
        \State $\bb$ $\leftarrow$ $\textbf{0\textsubscript{6x1}}$\\
        
        \For{$q$ $\in$ $\mathcal{T}_k$.getLeaves()}
        \For{$\mathcal{T}$ $\in$ $\mathcal{M}$}
        
        \State $^wq$ $\leftarrow$ $q$.transform($^w\bX_k$) \Comment{incremental}
        \State $l$ $\leftarrow$ $\mathcal{T}$.nearestNeighbor($^wq$)\Comment{data association}\\
        
        \State $e$, $\bJ$ $\leftarrow$ firstOrderApprox($q$, $l$) \Comment{system determination}
        \State $h$ $\leftarrow$ Huber($e$)
        \State $\bH$ $\leftarrow$ $\bH$ + $h$ $\bJ^T$$\bJ$
        \State $\bb$ $\leftarrow$ $\bb$ + $h$ $\bJ^T$$e$
        
        \EndFor
        \EndFor\\
        
        \State $^w\bX_k$ $\leftarrow$ solve($\bH$, $\bb$) \Comment{system resolution}
        \EndWhile\\
        
        \State \Return{$^w\bX_k$}	
    \end{algorithmic}
\end{algorithm}

\begin{figure}
  \centering
 \includegraphics[width=0.75\columnwidth]{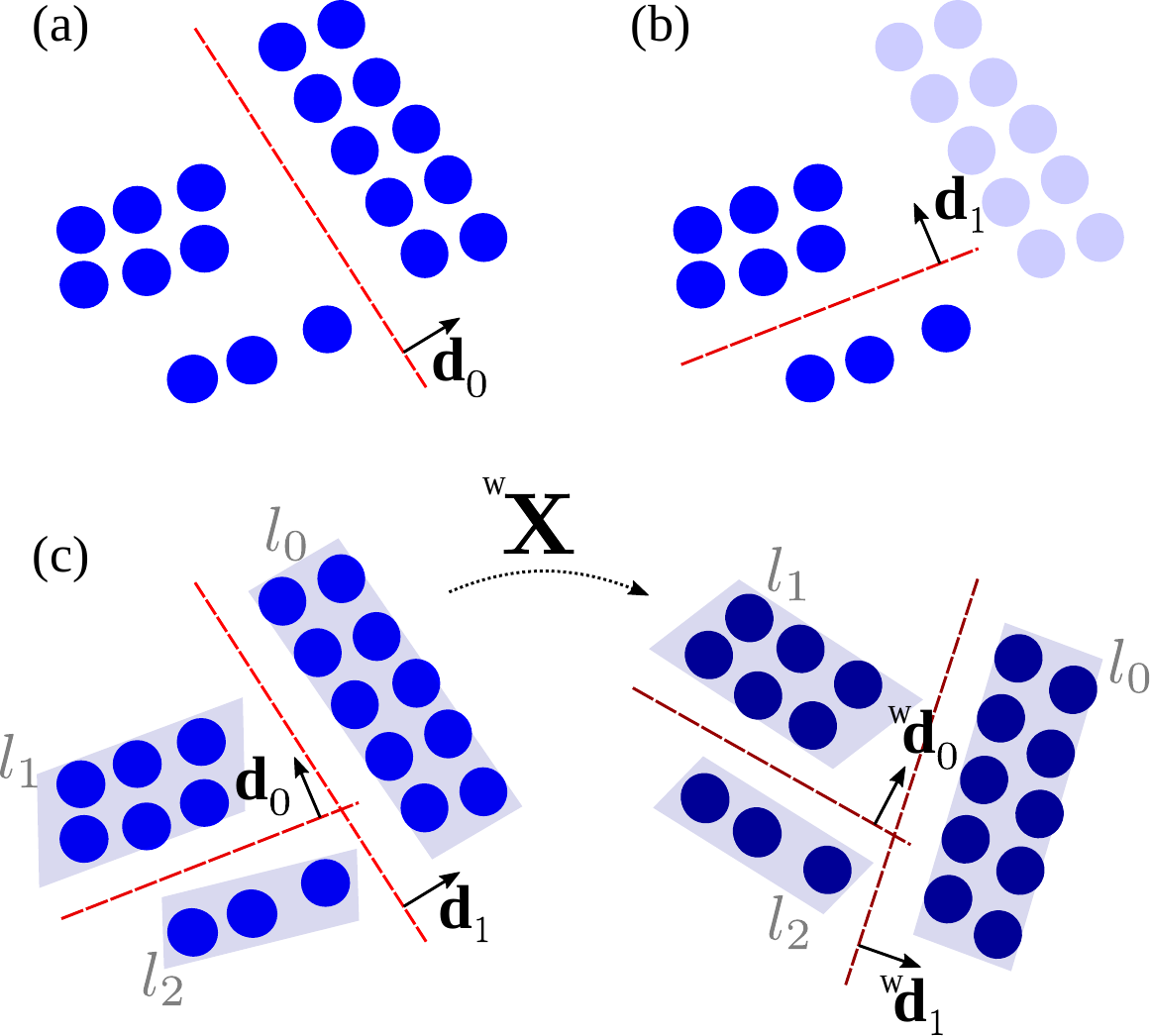}
  \caption{A 2D illustration of the kd-tree building and transformation processes. Our splitting criteria recursively divides the cloud with a plane whose normal is the direction $\bd$ of maximum spread (a, b), identified by \gls{pca}. The kd-tree is transformed in the world frame $w$ through an isometry $^w\bX$ (c). Our transformed kd-tree maintains the search properties.}
  \label{fig:kdtree}
\end{figure} 

\section{Runtimes}
\label{sec:runtimes}
\begin{figure}
  \centering
 \includegraphics[width=0.99\columnwidth]{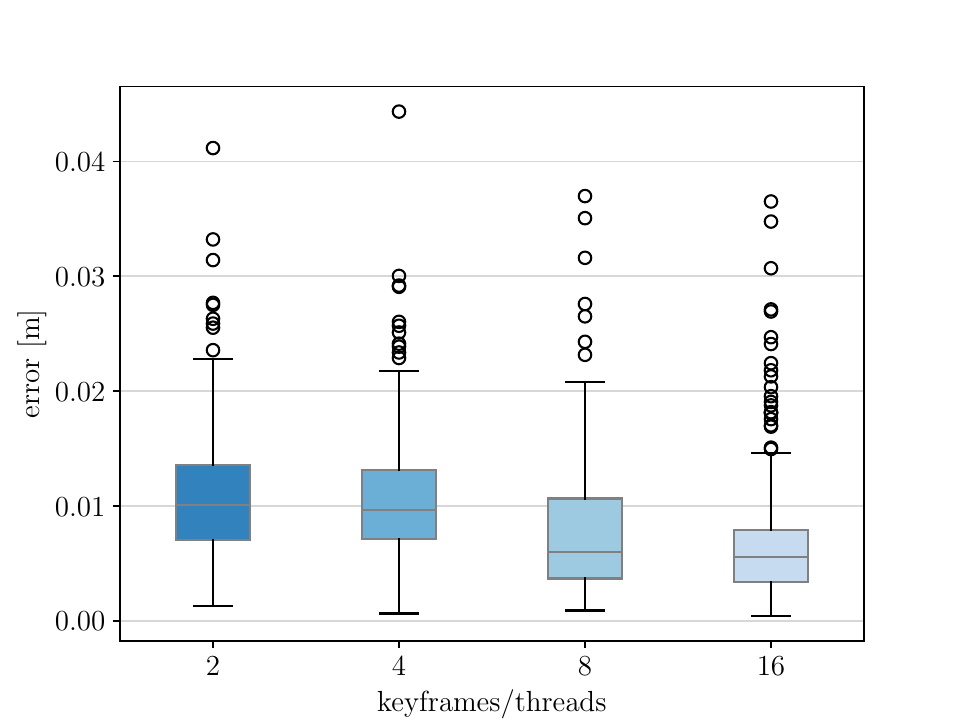}
  \caption{The image shows the RPE error in meters against the number of threads, thus the number of kd-trees in the forest/local map. We adopt this technique, in order to speed up computation while keeping good accuracy.}
  \label{fig:accvst}
\end{figure}

\begin{figure}
  \centering
 \includegraphics[width=0.99\columnwidth]{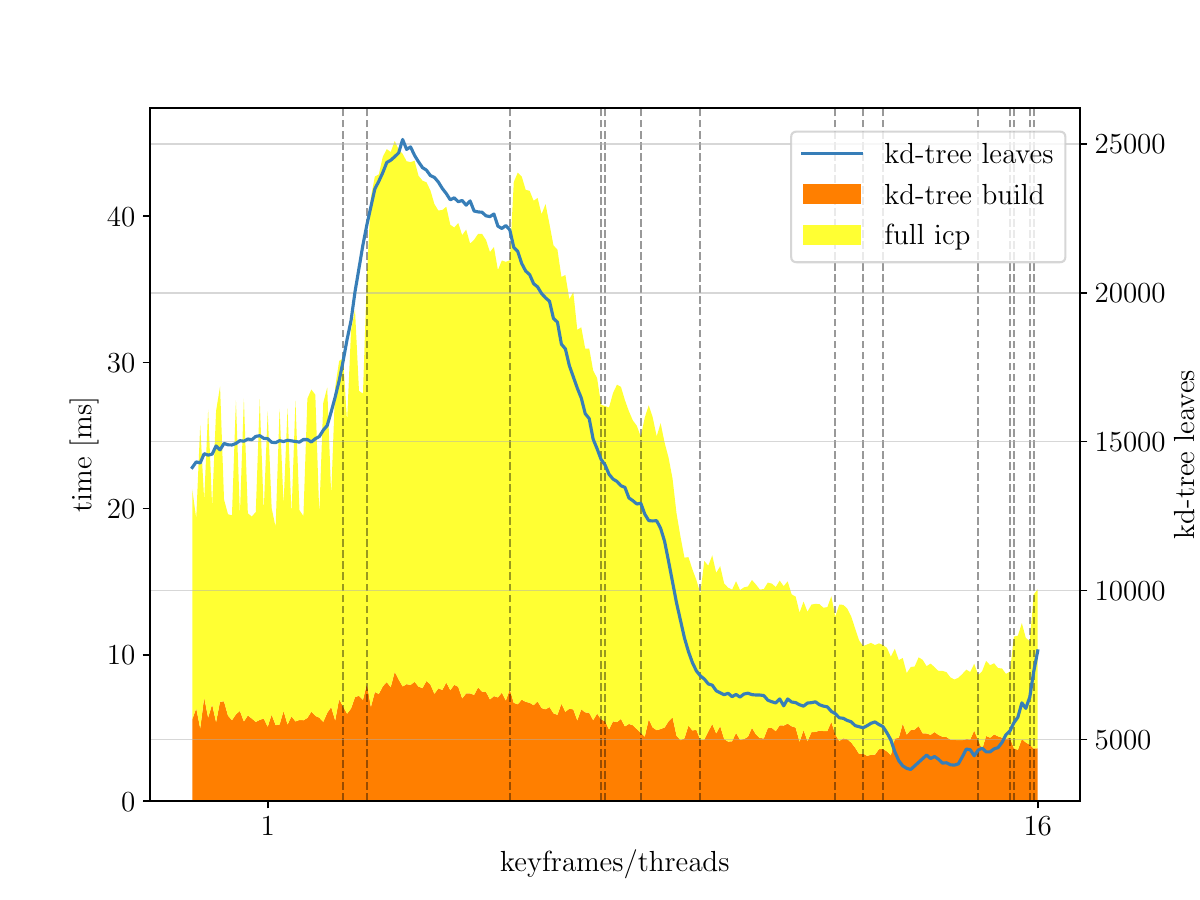}
  \caption{The image shows our runtimes in milliseconds for each point cloud. Specifically, we show kd-tree building and full registration process. The horizontal-axis highlights the number of keyframe from 1 to 16 (which is equivalent to the number of threads). Moreover, we show the correlation between ICP runtimes and number of leaves contained in the kd-trees.}
  \label{fig:profiling}
\end{figure}

Our implementation is \textit{anytime} because it can provide a valid estimate in real-time, scaling on the hardware. This means that we can return a valid solution even if computation is interrupted before it ends (\ie~less than maximum ICP iterations) to satisfy the real-time constraints. The algorithm is expected to find a more accurate solution the longer it keeps running. In order to speed up computation while maintaining good accuracy, we provide a multi-thread implementation, where each physical core manages a kd-tree of the local map $\cM$. Being our approach scan-to-model, enough amount of information needs to be encoded in the local map for higher accuracy. How the number of threads or kd-tree in the local map impacts the final result is showed in \figref{fig:accvst}.
In figure \figref{fig:profiling} instead, we show how our runtimes are related to the spatial extension of the observation (the farther the points of the point cloud the bigger the size of the observation). While this impacts the scan registration procedure, does not make substantial differences during kd-tree building. 

\section{Qualitative results}
\label{sec:qualitative}
\begin{figure}
  \centering
 \includegraphics[width=0.99\columnwidth]{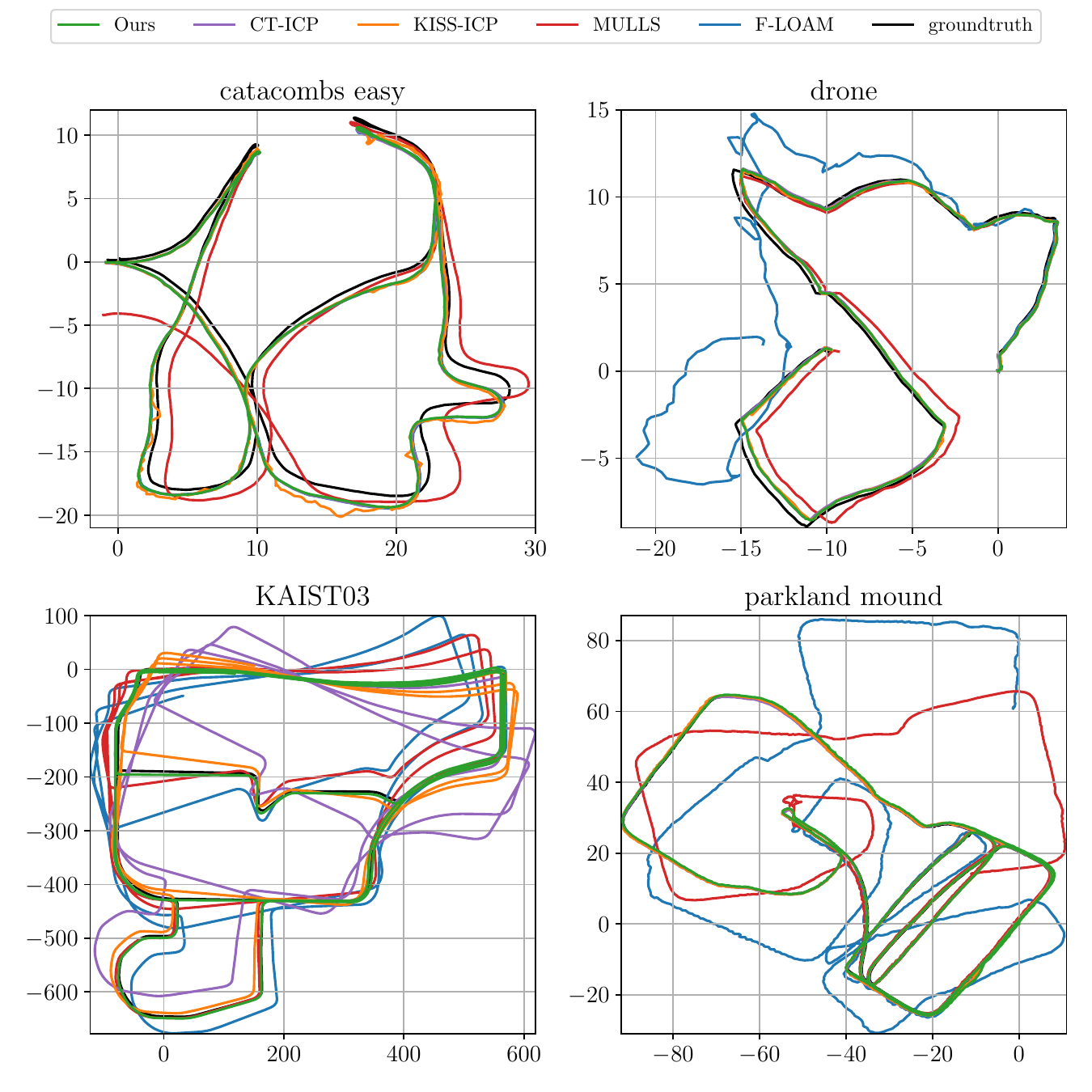}
  \caption{The plots illustrate the qualitative absolute consistency of the compared approaches within some sequences of the employed datasets.}
  \label{fig:plots_2}
\end{figure}

\begin{figure}
	\vspace{-1cm}
	\begin{subfigure}{0.38\textwidth}
		\hspace{2cm}
		\includegraphics[width=\linewidth]{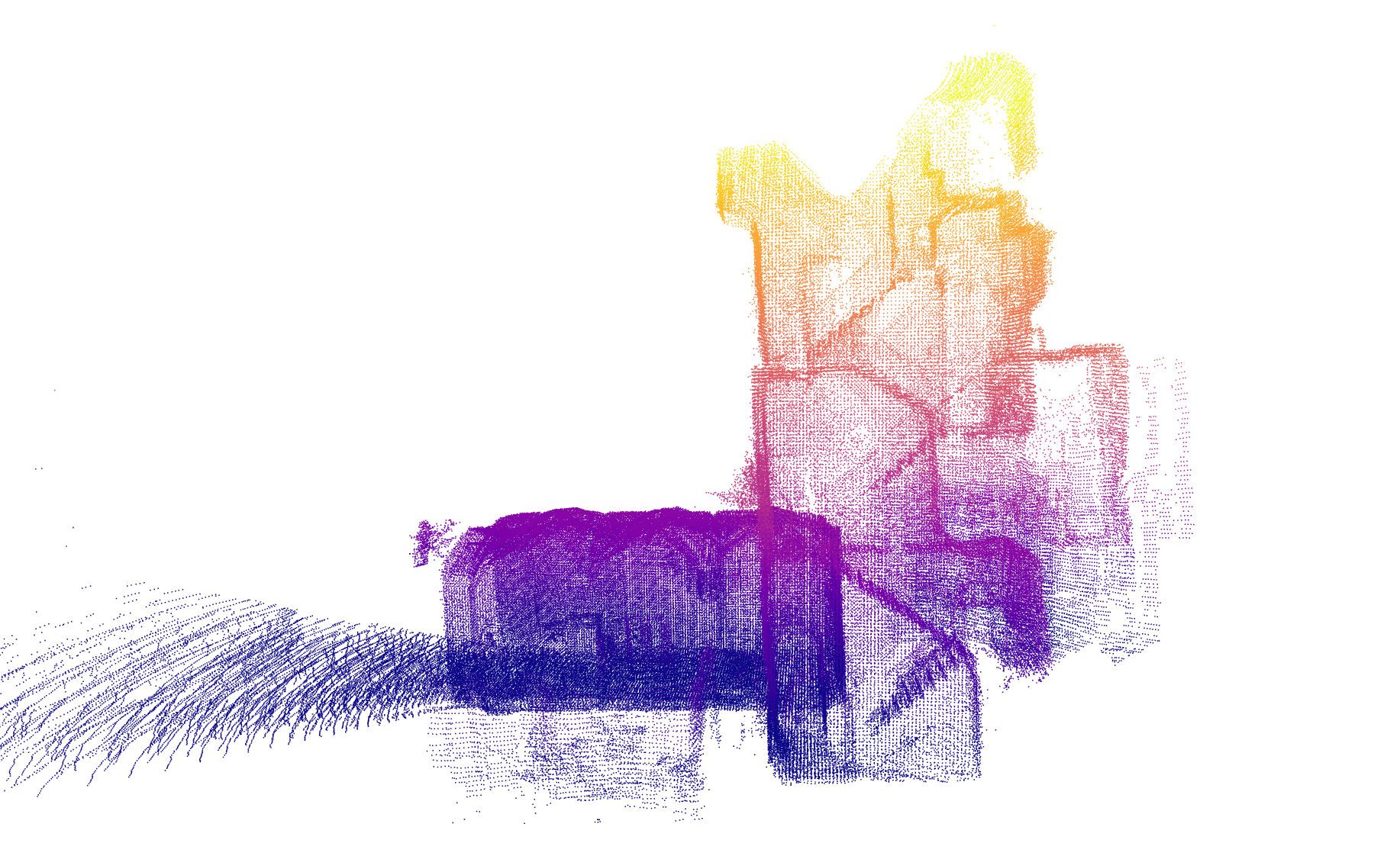}
		\caption{Ours}
		\label{fig:a3-stairs}
	\end{subfigure}
	\newline
	\begin{subfigure}{0.38\textwidth}
		\hspace{2cm}
		\includegraphics[width=\linewidth]{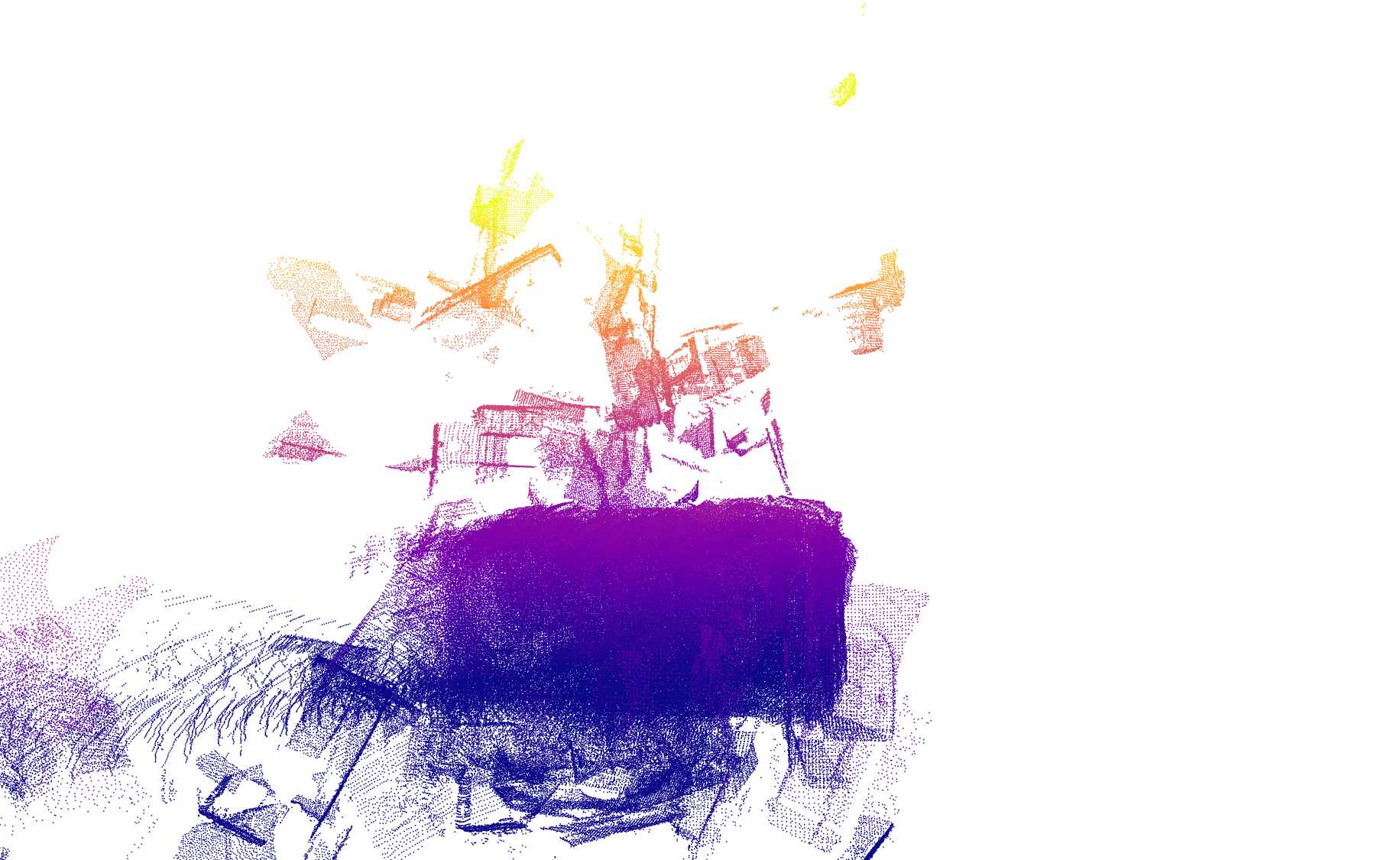}
		\caption{KISS-ICP \cite{vizzo2023kiss}}
		\label{fig:kiss-stairs}
	\end{subfigure}
	\newline
	\begin{subfigure}{0.38\textwidth}
		\hspace{2cm}
		\includegraphics[width=\linewidth]{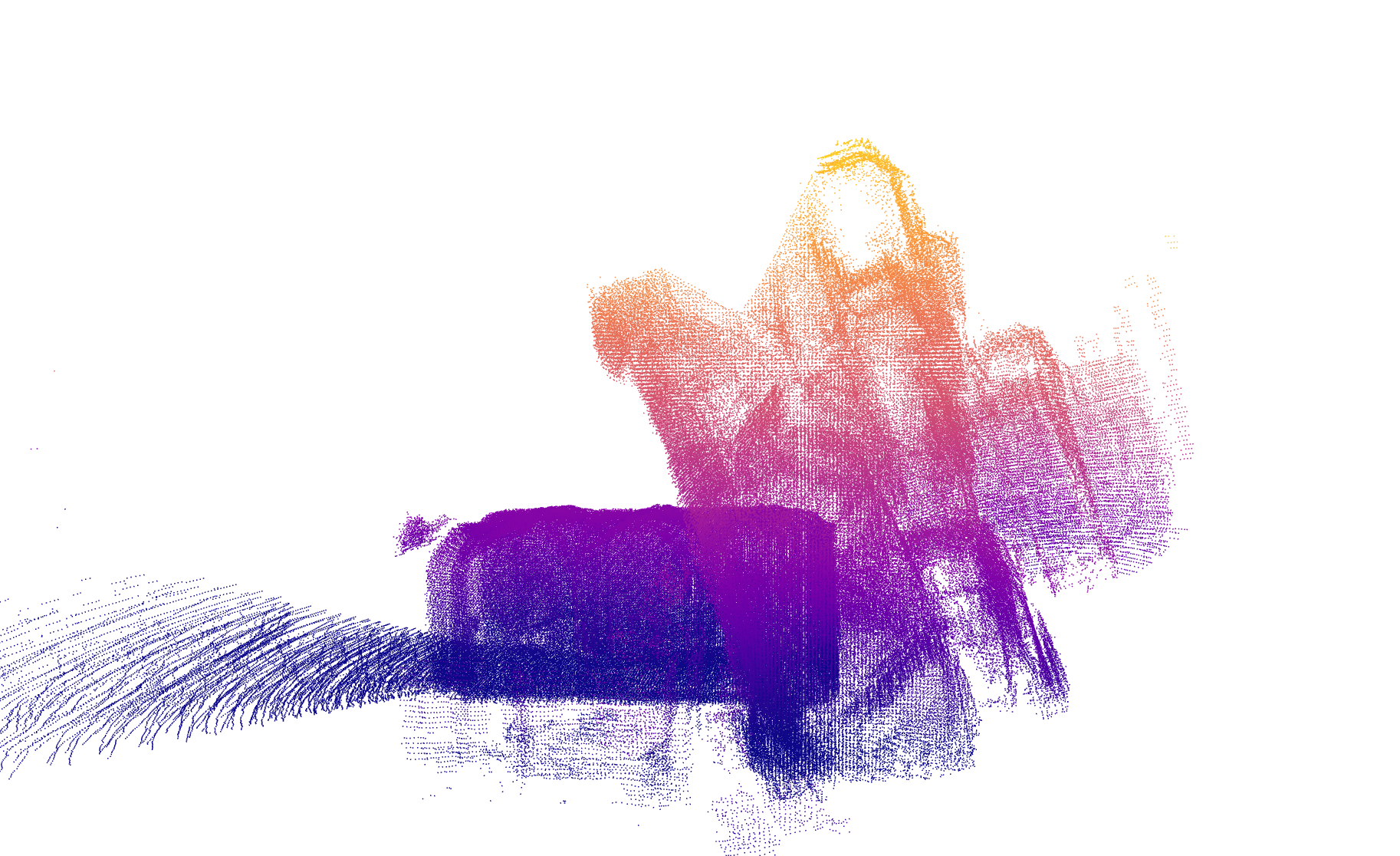}
		\caption{F-LOAM \cite{floam2016wang}}
		\label{fig:floam-stairs}
	\end{subfigure}
	\newline
	\begin{subfigure}{0.38\textwidth}
		\hspace{2cm}
		\includegraphics[width=\linewidth]{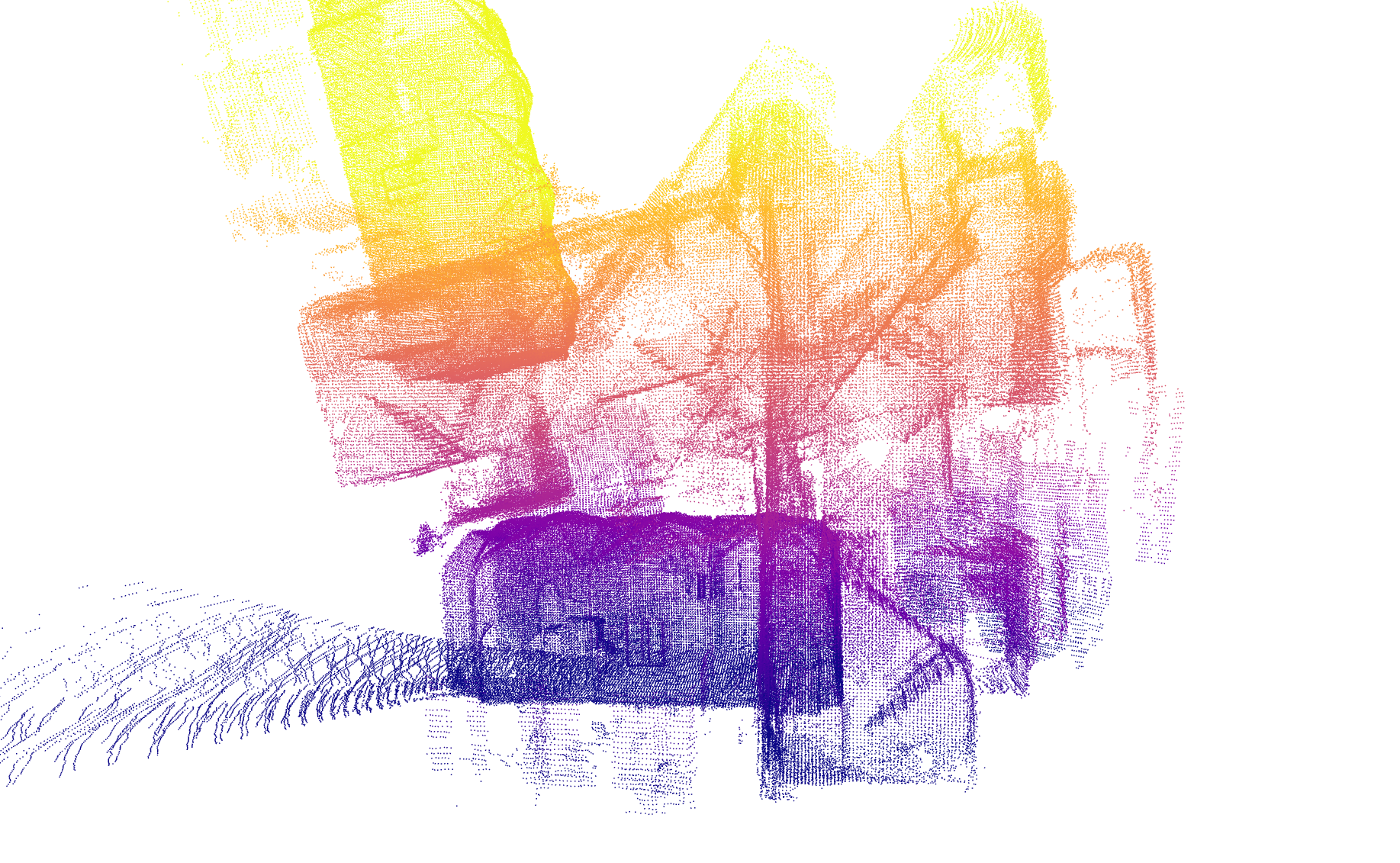}
		\caption{MULLS \cite{pan2021mulls}}
		\label{fig:mulls-stairs}
	\end{subfigure}
	\newline
	\begin{subfigure}{0.38\textwidth}
\hspace{2cm}
		\includegraphics[width=\linewidth]{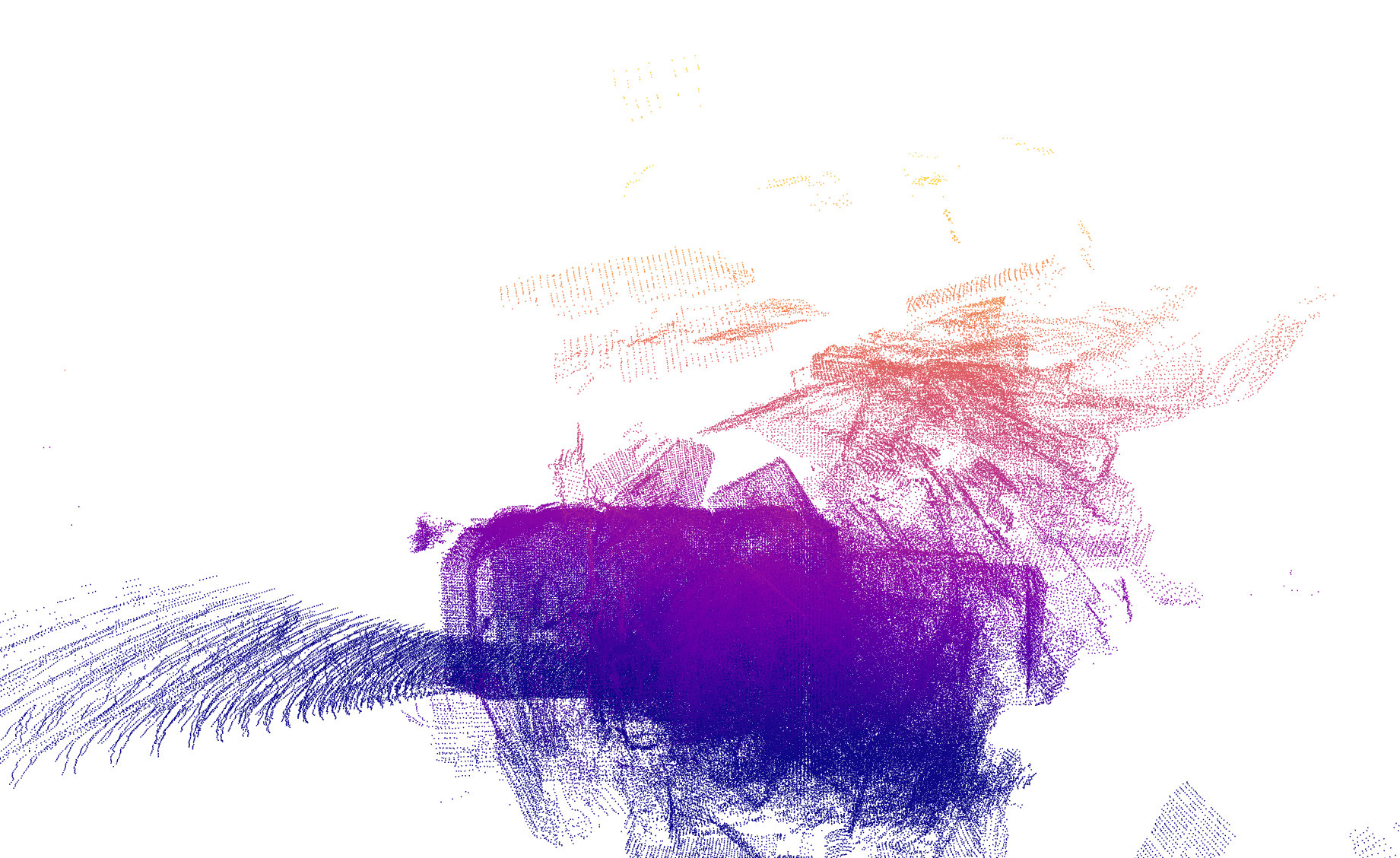}
		\caption{CT-ICP \cite{dellenbach2022ct}}
		\label{fig:ct-stairs}
	\end{subfigure}
	\caption{Qualitative results on the \textit{stairs} sequence of Newer College dataset \cite{ramezani2020newer}. Our approach is the only one that robustly handle vertical motion.}
	\label{fig:overall}
\end{figure}

\begin{figure*}
	\vspace{-17cm}
	\centering
	\begin{subfigure}{0.48\textwidth}
		\includegraphics[width=\linewidth]{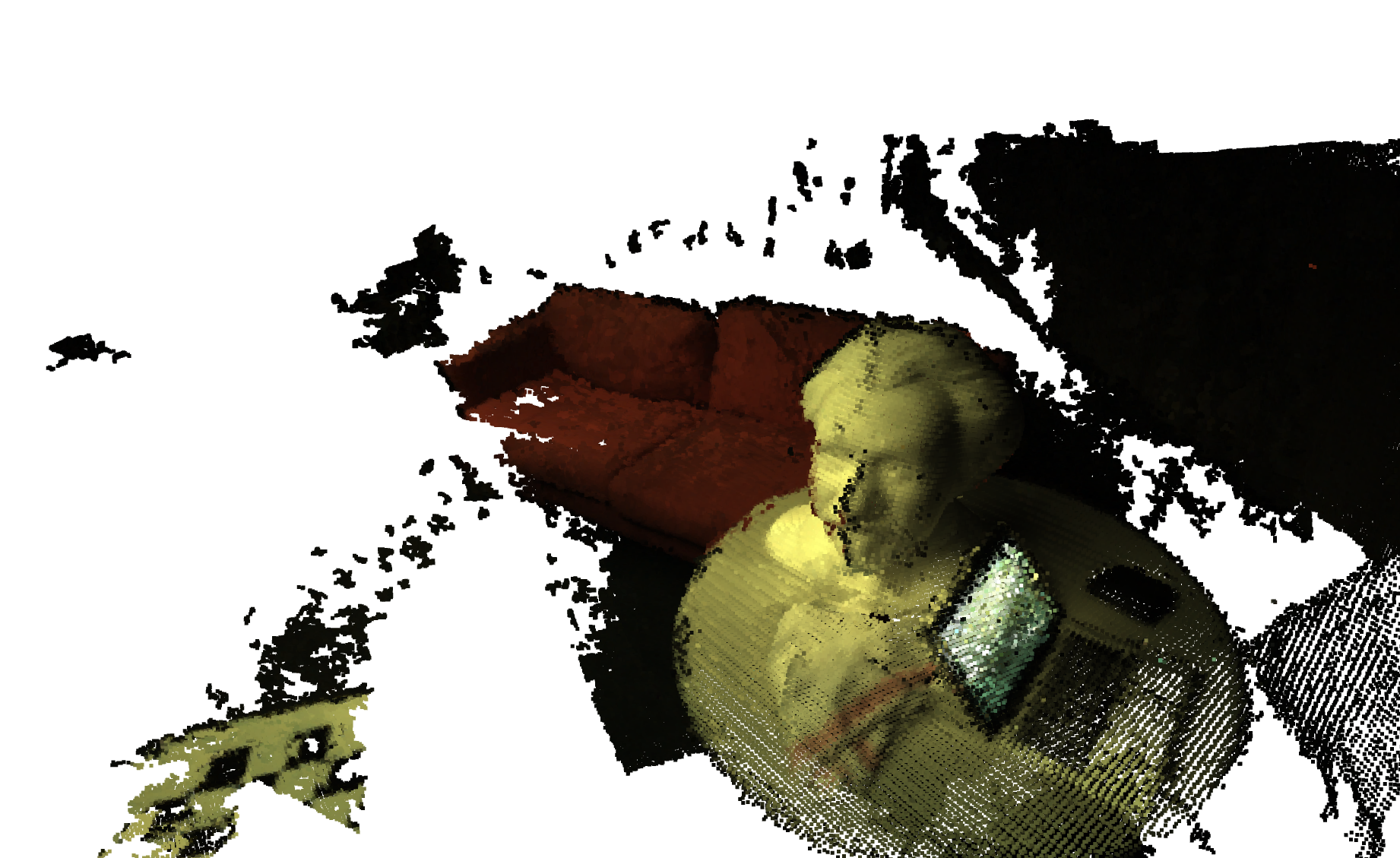}
		\caption{\textit{einstein\_1} sequence}
		\label{fig:einstein}
	\end{subfigure}
	\begin{subfigure}{0.48\textwidth}
		\includegraphics[width=\linewidth]{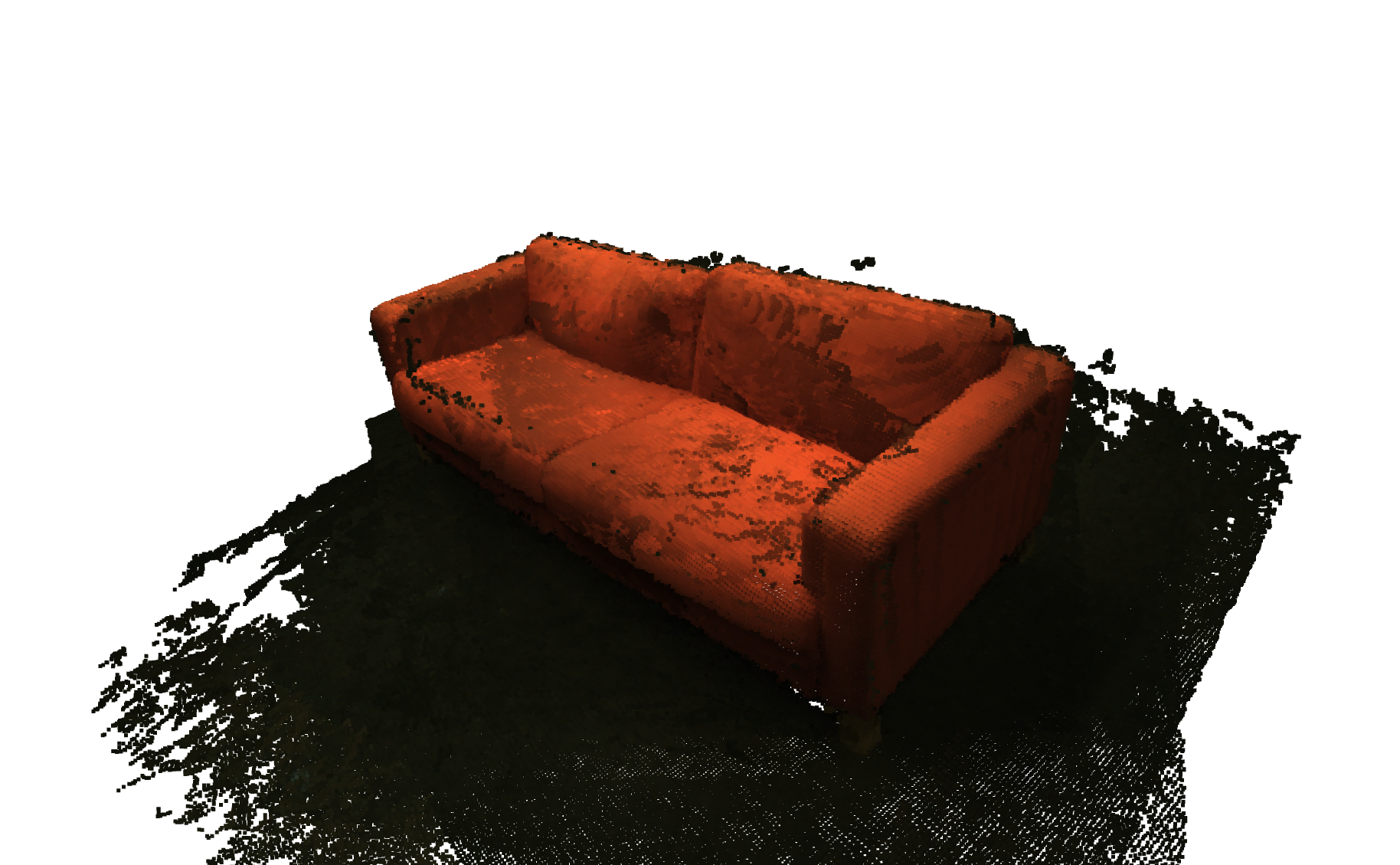}
		\caption{\textit{sofa\_1} sequence}
		\label{fig:sofa}
	\end{subfigure}
	\caption{Qualitative results of our pipeline on two sequences of ETH3D SLAM dataset \cite{schops2019bad}. Our approach is employable also with RGB-D clouds. Our registration schema does not refine any kind of structure. Something usually done with noisy RGB-D point clouds.}
	\label{fig:eth}
\end{figure*}

To complement our quantitative results, in \figref{fig:plots_2} we provide further plots for global consistency of compared trajectories. Moreover, in \figref{fig:overall} and \figref{fig:eth} we report qualitative plots for 3D mapping of the \textit{stairs} sequence and of two RGB-D sequences, respectively. \figref{fig:overall} shows that our pipeline is the only one employable in vertical environments, while \figref{fig:eth} demonstrates that our approach works also with point clouds generated with a different sensor.

\end{document}